\documentclass[11pt, a4paper, logo, copyright, nonumbering]{mll_style}
\usepackage{color}
\usepackage{epsfig}
\usepackage{graphicx}

\usepackage{booktabs}  %
\usepackage{tabularx}               %
\newcolumntype{C}{>{\centering\arraybackslash}X}
\usepackage{multirow}               %
\usepackage{diagbox}                %
\usepackage{hhline}                 %
\usepackage{color}                  %

\usepackage{booktabs}
\usepackage{extarrows}
\usepackage{makecell}
\usepackage{wrapfig}
\usepackage{colortbl}
\usepackage{tabularray}
\usepackage[most]{tcolorbox}
\tcbuselibrary{skins} %
\usepackage{tcolorbox}

\usepackage{amssymb}

\usepackage{adjustbox}
\usepackage{array}
\usepackage{floatflt}

\usepackage{bm}
\usepackage{nicefrac}
\usepackage{microtype}

\usepackage{changepage}
\usepackage{extramarks}
\usepackage{fancyhdr}
\usepackage{lastpage}
\usepackage{setspace}
\usepackage{soul}
\usepackage{xspace}

\usepackage{url}

\usepackage{enumerate}
\usepackage{enumitem}  %

\usepackage{makecell}

\usepackage{pifont} %

\usepackage{algorithm}
\usepackage{algpseudocode}
\usepackage{xcolor}

\usepackage[symbol]{footmisc}

\usepackage{scalefnt}

\usepackage{fontawesome5}

\usepackage{siunitx}

\usepackage{listings}

\newcolumntype{L}[1]{>{\raggedright\let\newline\\\arraybackslash\hspace{0pt}}m{#1}}
\newcolumntype{R}[1]{>{\raggedleft\let\newline\\\arraybackslash\hspace{0pt}}m{#1}}

\newcommand{\ignore}[1]{}

\makeatletter
\DeclareRobustCommand\onedot{\futurelet\@let@token\@onedot}
\def\@onedot{\ifx\@let@token.\else.\null\fi\xspace}

\makeatother

\definecolor{MyBlue}{rgb}{0.46, 0.50, 0.61}
\definecolor{MyDarkBlue}{rgb}{0,0.08,0.8}
\definecolor{MyDarkGreen}{RGB}{45,155,45}
\definecolor{MyDarkRed}{rgb}{0.8,0.02,0.02}
\definecolor{MyOrange}{rgb}{1.0, 0.4, 0.2}
\definecolor{MyPurple}{RGB}{111,0,255}
\definecolor{MyRed}{rgb}{0.8,0.0,0.0}
\definecolor{MyGold}{rgb}{0.75,0.6,0.12}
\definecolor{MyDarkgray}{rgb}{0.66, 0.66, 0.66}
\definecolor{MyBrown}{rgb}{0.65, 0.16, 0.16}
\definecolor{MyMutedRose}{rgb}{0.58, 0.29, 0.35}
\definecolor{JiayuanColor}{rgb}{0.60,0.43,0.48}
\definecolor{erranColor}{rgb}{24, 40, 113}

\definecolor{citecolor}{HTML}{696FAD}

\newif\ifpropositionfirstitem
\propositionfirstitemtrue

\definecolor{bggray}{HTML}{F5F5F5}
\definecolor{pvdblue}{HTML}{DAE8FC}
\definecolor{RoseQuartzBg}{HTML}{F7CAC9}
\definecolor{RoseQuartz}{HTML}{F5A798}
\definecolor{Serenity}{HTML}{92A8D1}
\definecolor{OrangeRed}{rgb}{1.0, 0.27, 0.0}
\definecolor{RoyalBlue}{cmyk}{1, 0.50, 0, 0}
\definecolor{Turquoise}{HTML}{0F4C81}
\definecolor{mint}{rgb}{0.24, 0.71, 0.54}
\definecolor{green}{rgb}{0.0, 0.120, 0.0}

\newdimen\abovecrulesep
\newdimen\belowcrulesep
\makeatletter
\patchcmd{\@@@cmidrule}{\aboverulesep}{\abovecrulesep}{}{}
\patchcmd{\@xcmidrule}{\belowrulesep}{\belowcrulesep}{}{}
\makeatother

\definecolor{mybluetitle}{HTML}{4B527E} %

\definecolor{mygreen}{RGB}{0,150,0}
\definecolor{boxbackground}{HTML}{F0F7FF}  %
\definecolor{boxborder}{HTML}{D0D9E5}      %
\definecolor{accentblue}{HTML}{4A86E8}     %
\definecolor{lightblue}{HTML}{EEF3FF}  %
\definecolor{bordergray}{HTML}{CCCCCC}  %
\definecolor{headerblue}{HTML}{2C5AA0}  %

\definecolor{lavenderframe}{HTML}{E6E6FA}  %
\definecolor{lighterlav}{HTML}{F5F5FF}  %
\definecolor{codegray}{rgb}{0.5,0.5,0.5}  %
\definecolor{codepurple}{HTML}{483D8B}  %
\definecolor{backcolour}{HTML}{F5F5FF}  %
\lstdefinestyle{mystyle}{
    backgroundcolor=\color{backcolour},
    commentstyle=\color{headerblue},
    keywordstyle=\color{codepurple},
    numberstyle=\tiny\color{codegray},
    stringstyle=\color{codepurple},
    basicstyle=\ttfamily\scriptsize,
    breakatwhitespace=false,
    breaklines=true,
    captionpos=b,
    keepspaces=true,
    frame=none,
    numbersep=5pt,
    showspaces=false,
    showstringspaces=false,
    showtabs=false,
    tabsize=2
}

\definecolor{jsonkey}{RGB}{44, 130, 201}     %
\definecolor{jsonstring}{RGB}{255, 140, 0}   %
\definecolor{jsonnumber}{RGB}{34, 139, 34}   %

\lstdefinelanguage{json}{
    basicstyle=\ttfamily\small,
    numbers=left,
    numberstyle=\tiny\color{gray},
    stepnumber=1,
    numbersep=5pt,
    showstringspaces=false,
    breaklines=true,
    frame=none,
    backgroundcolor=\color{gray!5},
    literate=
     *{:}{{{\color{jsonkey}:}}}{1}
      {,}{{{\color{jsonkey},}}}{1}
      {"}{{{\color{jsonstring}"}}}{1}
      {[}{{{\color{jsonkey}[}}}{1}
      {]}{{{\color{jsonkey}]}}}{1}
      {0}{{{\color{jsonnumber}0}}}{1}
      {1}{{{\color{jsonnumber}1}}}{1}
      {2}{{{\color{jsonnumber}2}}}{1}
      {3}{{{\color{jsonnumber}3}}}{1}
      {4}{{{\color{jsonnumber}4}}}{1}
      {5}{{{\color{jsonnumber}5}}}{1}
      {6}{{{\color{jsonnumber}6}}}{1}
      {7}{{{\color{jsonnumber}7}}}{1}
      {8}{{{\color{jsonnumber}8}}}{1}
      {9}{{{\color{jsonnumber}9}}}{1}
}

\newtcblisting{jsonbox}{
  listing engine=listings,
  colback=gray!3!white,
  colframe=gray!75!black,
  boxrule=0.4mm,
  arc=2mm,
  outer arc=2mm,
  breakable,
  enhanced,
  listing only,
  listing options={language=json}
}

\newtcolorbox{promptbox}[2][]{ %
    enhanced,
    breakable,
    boxsep=5pt,
    left=9pt,
    right=7pt,
    top=5pt,
    bottom=5pt,
    colback=boxbackground,
    colframe=boxborder,
    boxrule=0.5pt,
    arc=4pt,
    frame hidden, %
    borderline west={3pt}{0pt}{accentblue},
    shadow={0.5pt}{0.5pt}{1.5pt}{black!10},
    fontupper=\normalsize,
    title=#2, %
    colbacktitle=accentblue, %
    coltitle=white,         %
    fonttitle={\fontsize{9}{11}\selectfont\bfseries}, %
    attach boxed title to top left={yshift=-2.5mm, xshift=3.2mm},
    boxed title style={
        enhanced,
        left=3pt,
        right=3pt,
        top=1pt,    %
        bottom=1pt, %
        boxsep=2pt,
        arc=3pt,
        boxrule=0pt,
        colback=accentblue,
    },
    #1 %
}

\newtcolorbox{notitlepromptbox}[1][]{
    enhanced,
    breakable,
    boxsep=5pt,          %
    left=9pt,            %
    right=7pt,           %
    top=5pt,             %
    bottom=5pt,          %
    colback=boxbackground,
    colframe=boxborder,
    boxrule=0.5pt,
    arc=4pt,             %
    frame hidden,
    borderline west={3pt}{0pt}{accentblue},  %
    shadow={0.5pt}{0.5pt}{1.5pt}{black!10},  %
    notitle,
    fontupper=\normalsize,    %
    #1
}

\newtcolorbox{onebox}[2][]{
    enhanced, 
    center title,
    left*=0pt, right*=0pt,
    boxsep=2pt, left=5pt, right=5pt,
    skin first=enhanced,
    skin middle=enhanced,
    skin last=enhanced,
    colframe = mybluetitle!90,
  colback  = mybluetitle!10,
    fonttitle=\bfseries\rmfamily\fontfamily{phv}\selectfont,
    title={\footnotesize\strut{#2}  \refstepcounter{subsubsection} \addcontentsline{toc}{subsubsection}{\string\numberline{\thesubsubsection}#2}
    },
    #1
    }

\usepackage{algorithm}          
\usepackage{algpseudocode}      
\usepackage{amsmath}            
\usepackage{amsfonts}           
\usepackage{bm}   

\usepackage{amsfonts}
\usepackage[numbers]{natbib}
\usepackage{graphicx}
\usepackage{xspace}
\usepackage{xcolor}
\definecolor{highlightgray}{RGB}{220, 220, 220}
\usepackage{adjustbox}
\usepackage{array}
\usepackage{float}
\usepackage{geometry}
\usepackage{dblfloatfix}
\usepackage{enumitem}
\usepackage{setspace}
\usepackage{booktabs}
\usepackage{multirow}
\usepackage{tabularx}
\usepackage{siunitx}
\usepackage{amsmath,array}

\usepackage{pdflscape}

\usepackage{amsmath}
\usepackage{amssymb}
\usepackage{amsfonts}
\usepackage{mathtools}
\usepackage{bm}
\usepackage{dsfont}

\usepackage{tikz}
\usepackage{pgfplots}
\pgfplotsset{compat=1.18}
\usepackage{algorithm}
\usepackage{algpseudocode}
\usepackage{listings}

\usepackage{cleveref}
\hypersetup{colorlinks=true}

\usepackage{enumitem}
\usepackage{todonotes}
\usepackage{fontawesome5}
\usepackage[most]{tcolorbox}
\usepackage{CJKutf8}
\usepackage{lipsum}

\usepackage[misc]{ifsym}
\usepackage[table]{xcolor}
\usepackage{makecell}
\usepackage{colortbl}
\usepackage{bbm}
\usepackage{wrapfig}

\definecolor{kbBlue}{RGB}{92, 131, 227}     
\definecolor{kbLight}{RGB}{237, 244, 255}   
\definecolor{kbBar}{RGB}{92, 131, 227}      
\usepackage{tcolorbox}
\tcbuselibrary{skins,breakable}

\newtcolorbox{prompt}{%
  enhanced,
  breakable,
  colback=kbLight,
  colframe=kbLight,
  boxrule=0pt,
  arc=8pt,
  left=12pt, right=12pt, top=12pt, bottom=12pt,
  borderline west={4pt}{0pt}{kbBar},
}
\newcommand{\myrot}[2][65]{\rotatebox{#1}{\scriptsize\strut #2}}
\newcommand{\myrothead}[2][65]{\makebox[0pt][c]{\myrot[#1]{#2}}}

\algdef{SE}[FORALL]{ForAll}{EndForAll}[1]{\algorithmicforall\ #1\ }{}
\algtext*{EndForAll}

\definecolor{tableblue}{RGB}{201,226,239}

\makeatletter
\def\@BTrule[#1]{%
  \ifx\longtable\undefined
    \let\@BTswitch\@BTnormal
  \else\ifx\hline\LT@hline
    \nobreak
    \let\@BTswitch\@BLTrule
  \else
     \let\@BTswitch\@BTnormal
  \fi\fi
  \global\@thisrulewidth=#1\relax
  \ifnum\@thisruleclass=\tw@\vskip\@aboverulesep\else
  \ifnum\@lastruleclass=\z@\vskip\@aboverulesep\else
  \ifnum\@lastruleclass=\@ne\vskip\doublerulesep\fi\fi\fi
  \@BTswitch}
\makeatother

\addto\extrasenglish{
}

 {\begin{list}{}%
         {\setlength{\leftmargin}{#1}}%
         \item[]%
 }
 {\end{list}}

\usepackage[toc,page,header]{appendix}
\usepackage{minitoc}

\reportnumber{001} %

\title{\centering Stepping VLMs onto the Court: Benchmarking Spatial Intelligence in Sports
}

\date{}

\author[*]{
Yuchen Yang$^{1,2,*}$,
Yuqing Shao$^{4,2,*}$,
Duxiu Huang$^{5,*}$,
Linfeng Dong$^{6,2,*}$,
Yifei Liu$^{7,2}$,
Suixin Tang$^{4}$,
Xiang Zhou$^{4}$,
Yuanyuan Gao$^{8,2}$,
Wei Wang$^{2}$,
Yue Zhou$^{9}$,
Xue Yang$^{3}$,
Yanfeng Wang$^{3}$,
Xiao Sun$^{2}$,
Zhihang Zhong$^{3,}\textsuperscript{\Letter}$

{\small $^*$Equal Contribution; $^{\textsuperscript{\Letter}}$Corresponding Authors}
\\
\small \textsuperscript{1}Fudan~University,
\textsuperscript{2}Shanghai~Artificial~Intelligence~Laboratory,
\textsuperscript{3}Shanghai~Jiao~Tong~University,
\textsuperscript{4}East~China~University~of~Science~and~Technology,
\textsuperscript{5}Southeast~University,
\textsuperscript{6}Zhejiang~University,
\textsuperscript{7}Beihang~University,
\textsuperscript{8}Hong~Kong~University~of~Science~and~Technology,
\textsuperscript{9}East~China~Normal~University
\\
\vspace{-10pt}
}

\begin{abstract}
Sports have long attracted broad attention as they push the limits of human physical and cognitive capabilities. 
Amid growing interest in spatial intelligence for vision-language models (VLMs), sports provide a natural testbed for understanding high-intensity human motion and dynamic object interactions.
To this end, we present \textbf{CourtSI}, the first large-scale spatial intelligence dataset tailored to sports scenarios.
CourtSI contains over 1M QA pairs, organized under a holistic taxonomy that systematically covers spatial counting, distance measurement, localization, and relational reasoning, across representative net sports including badminton, tennis, and table tennis.
Leveraging well-defined court geometry as metric anchors, we develop a semi-automatic data engine to reconstruct sports scenes, enabling scalable curation of CourtSI.
In addition, we introduce \textbf{CourtSI-Bench}, a high-quality evaluation benchmark comprising 3,686 QA pairs with rigorous human verification.
We evaluate 25 proprietary and open-source VLMs on CourtSI-Bench, revealing a remaining human–AI performance gap and limited generalization from existing spatial intelligence benchmarks. These findings indicate that sports scenarios expose limitations in spatial intelligence capabilities captured by existing benchmarks.
Further, fine-tuning Qwen3-VL-8B on CourtSI improves accuracy on CourtSI-Bench by \textbf{23.5} percentage points. The adapted model also generalizes effectively to \textbf{CourtSI-Ext}, an evaluation set built on a similar but unseen sport, and demonstrates enhanced spatial-aware commentary generation. 
Together, these findings demonstrate that CourtSI provides a scalable pathway toward advancing spatial intelligence of VLMs in sports.

\vspace{8pt}

\textbf{Website}: \href{https://visionary-laboratory.github.io/CourtSI}{https://visionary-laboratory.github.io/CourtSI}\\~
\textbf{Code}: \href{https://github.com/Visionary-Laboratory/CourtSI}{https://github.com/Visionary-Laboratory/CourtSI}\\~
\textbf{Email: } \texttt{zhongzhihang95@gmail.com}

\end{abstract}

\begin{document}
\begin{CJK*}{UTF8}{gbsn}

\doparttoc %
\faketableofcontents %

\maketitle

\section{Introduction}

As Vision-Language Models (VLMs) continue to achieve strong performance in semantic understanding and 2D visual reasoning, researchers have begun to explore VLMs' ability to perceive and reason about the 3D world.
This shift has led to the emergence of \textit{spatial intelligence}~\cite{chen2024spatialvlm} as a focused research direction, aiming to equip models with foundational capabilities required for effective interaction with the physical world in the pursuit of AGI.

Current efforts~\cite{VSI,MMSI,SPAR,SpatialRGPT,Sensenova,MindCube} primarily focus on boosting the spatial understanding of modern VLMs, along with developing diverse benchmarks for evaluation across multiple spatial dimensions.
However, the datasets proposed in these works concentrate on static scenes and rigid objects, resulting in a relatively narrow coverage of spatial subjects.
In contrast, humans, critical subjects in real-world environments characterized by non-rigid deformations and articulated body constraints, remain underexplored.
Sports scenarios, characterized by high-intensity human motion and dynamic object interactions, provide a natural but challenging testbed for investigating spatial intelligence at a fine-grained level.

\begin{figure}[t]
    \centering
    \includegraphics[width=\linewidth]{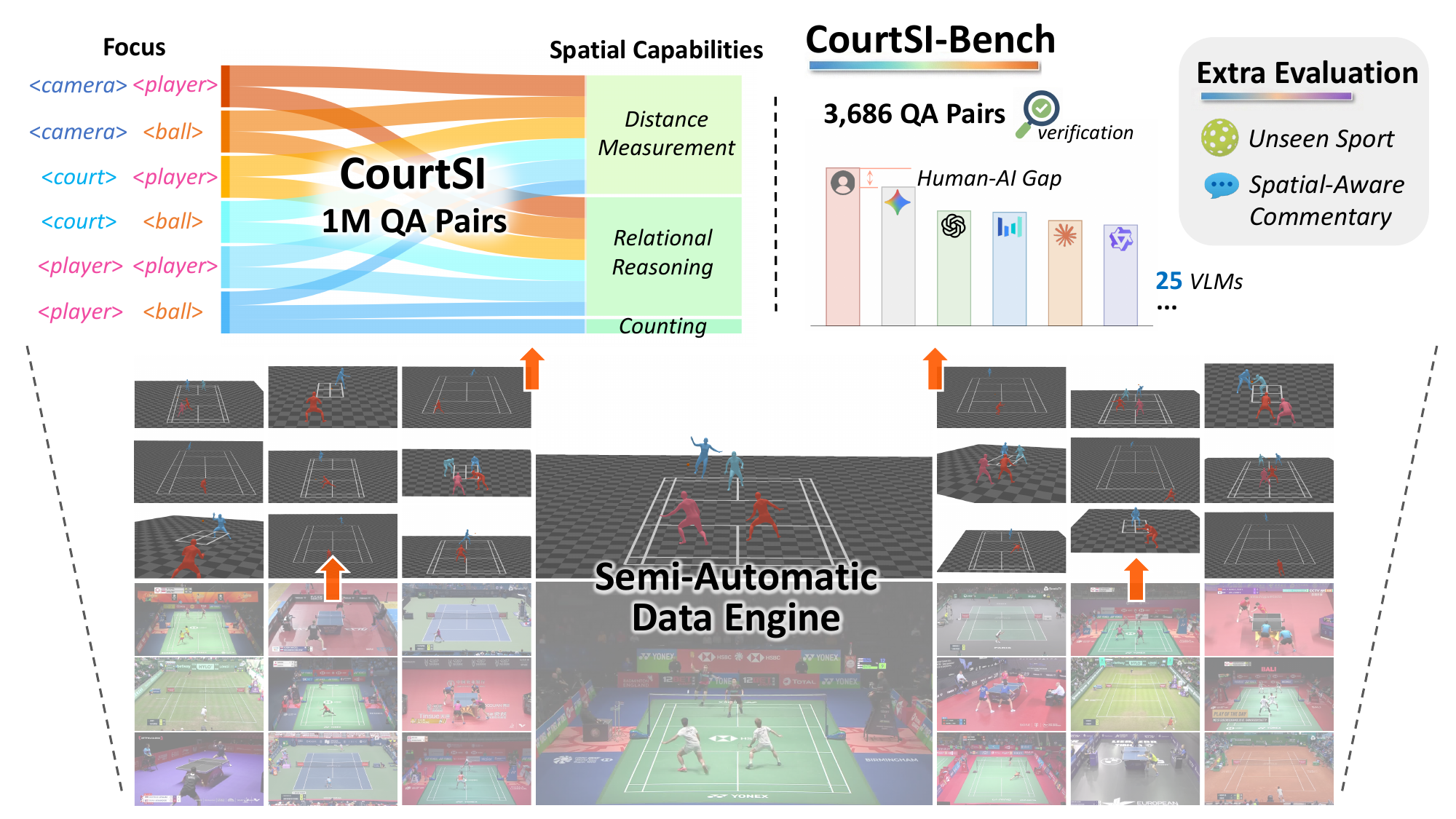}
    \caption{Overview. We introduce a semi-automatic data engine that reconstructs sports scenes in 3D with court, player, and ball locations. Built upon this pipeline, we present CourtSI and CourtSI-Bench, the first large-scale spatial intelligence dataset and benchmark for sports scenarios. In addition, we provide extra evaluation protocols to validate applicability on an unseen sport and spatial-aware commentary.}
    \label{fig:teaser}
\end{figure}

Motivated by the nature of sports scenarios, as illustrated in \cref{fig:teaser}, we present CourtSI and CourtSI-Bench, the first large-scale dataset and benchmark dedicated to spatial intelligence in sports.
Our work introduces sports as a new and challenging scenario for spatial intelligence, while simultaneously extending existing VLM benchmarks for sports understanding~\cite{xia2024sportu,xia2025sportr,he2025finebadminton,gao2025fsbench} beyond activity-centric to fine-grained spatial reasoning.

To obtain data at scale, we design a semi-automatic reconstruction data engine that recovers 3D scene information from monocular images.
Unlike general in-the-wild environments, sports courts provide well-defined geometric structures with fixed metric scales. 
Leveraging this property, we jointly optimize camera intrinsics and extrinsics from court corner correspondences using a Perspective-n-Point (PnP) solver, thereby establishing a unified world coordinate system anchored to the court geometry.
Locating players and balls into this geometry-aligned space ensures consistent and physically grounded spatial reasoning across scenes.
Specifically, for players, we adopt PromptHMR~\cite{phmr} to recover human meshes in the SMPL-X~\cite{smplx} representation within the camera coordinate system, capturing fine-grained pose and shape information.
Ball positions in the images are manually annotated.
We observe that existing monocular depth estimation methods fail to produce reliable metric reconstruction.
Instead, we manually estimate object heights relative to the court plane to enable accurate camera-to-world transformation.
With strict 3D quality control over a multi-view set, our pipeline achieves $cm$-level accuracy, providing a reliable foundation for subsequent data curation.

Building upon the reconstruction engine, we construct CourtSI by converting 3D sports states into large-scale question-answer (QA) pairs under a holistic taxonomy.
Specifically, we filter data from the well-organized sports dataset, RacketVision~\cite{dong2025racketvision}, which includes badminton, tennis, and table tennis in broadcast views.
The camera viewpoints in broadcast footage mitigate unnecessary viewpoint variance, allowing models to focus on learning spatial relationships.
We design QA templates that systematically cover (i) \emph{spatial counting}, (ii) \emph{distance measurement}, (iii) \emph{localization}, and (iv) \emph{relational reasoning}, instantiated over players, balls, and the court. 
Answers are automatically derived from the reconstructed 3D states, resulting in over 1M QA pairs for training.
To enable rigorous evaluation, we further curate CourtSI-Bench, comprising 3,686 high-quality QA pairs with careful human verification.

We comprehensively evaluate 25 state-of-the-art proprietary and open-source VLMs on CourtSI-Bench. 
Even the strongest baseline remains a gap behind humans, particularly on distance measurement tasks.
Furthermore, models trained on existing spatial intelligence benchmarks generalize poorly to CourtSI-Bench, suggesting that current datasets fail to sufficiently capture the challenges posed by dynamic sports scenarios.
To assess the training utility of CourtSI, we conduct supervised fine-tuning of Qwen3-VL-8B~\cite{qwen3vl}, improving accuracy by $23.5$ percentage points on CourtSI-Bench, with particularly significant gains in distance measurement.
To expand the evaluation, we introduce CourtSI-Ext, a benchmark constructed from pickleball, a similar yet unseen net sport. The fine-tuned model demonstrates strong generalization to this new sport, indicating that CourtSI fosters transferable spatial reasoning capabilities.
Additionally, we explore spatial-aware commentary generation by prompting VLMs to incorporate spatial relationships into commentary for CourtSI-Bench samples.
User studies demonstrate improved spatial understanding while preserving overall linguistic quality after fine-tuning on CourtSI.
Collectively, these results validate the effectiveness of CourtSI and highlight its potential as a scalable pathway toward advancing spatial intelligence of VLMs in sports.

Our contributions are summarized in threefold:

\begin{itemize}
    \item We introduce CourtSI and CourtSI-Bench, the first large-scale spatial intelligence dataset and benchmark in sports, establishing a testbed for fine-grained, human-centric spatial reasoning beyond static object-centric datasets.
    \item We develop a semi-automatic data engine that recovers accurate 3D scene states from broadcast net sports, enabling scalable data curation.
    \item We conduct a comprehensive evaluation of 25 state-of-the-art VLMs and examine the impact of fine-tuning, along with cross-sport generalization on CourtSI-Ext and spatial-aware commentary generation.
\end{itemize}
\section{Related Work}

\subsection{Spatial Intelligence of VLMs}
Along with the development of Vision-Language Models (VLMs), researchers have increasingly questioned their ability to reason about relationships of perceived 3D objects when trained primarily on web-scale data on the image plane~\cite{chen2024spatialvlm}.
This limitation has motivated the emergence of spatial intelligence, a term used to characterize models’ capabilities in 3D spatial reasoning.
Such capabilities are widely regarded as the foundation of reliable interaction with the physical world, in the broader pursuit of general intelligence~\cite{du2024embspatial,team2025gemini,song2025robospatial}.

To better characterize and advance these capabilities, the research community has developed both dedicated benchmarks and specialized approaches.
From a benchmarking perspective,
VSI~\cite{VSI} collects data with camera browsing inside indoor environments, requiring models to perceive, memorize, and recall spatial layouts.
Subsequent works extend evaluation across different dimensions of spatial understanding~\cite{wu2025spatialscore,deng2025internspatial,SITE,MMSI,mmsi-video}. MindCube~\cite{MindCube} focuses on sparse-view reasoning, while ViewSpatial~\cite{li2025viewspatial} emphasizes allocentric spatial reasoning. The underlying data sources have also expanded from structured indoor datasets such as ScanNet~\cite{dai2017scannet} to more diverse and less constrained 3D collections~\cite{ling2024dl3dv,sun2025spacevista}.
From a methodological perspective, a common approach is to enhance spatial intelligence through supervised fine-tuning or reinforcement learning strategies~\cite{li2026spatialladder,cai2026depthlm,ouyang2025spacer,yang2025visual,yang2025cambrian,Sensenova,gao2026holispatial}.
In addition, several works~\cite{SpatialRGPT,daxberger2025mm,zhang2026on,wu2025spatial,zheng2025learning} improve spatial reasoning by modifying the visual backbone, incorporating stronger geometric priors.
In contrast, our work focuses on sports scenarios, with a particular emphasis on human-centric spatial reasoning.

\subsection{Sport Understanding}
Sports understanding has long been an active research area, encompassing tasks such as action recognition~\cite{giancola2018soccernet,ibrahim2016hierarchical,yang2025sga} and analysis~\cite{wang2024tacticai,dong2024lucidaction,rao2025towards}.
The advent of language models has substantially accelerated progress in this domain via stronger end-to-end reasoning capability, especially in captioning and commentary generation~\cite{mkhallati2023soccernet,xi2025simple,xi2025player}. 
More recently, unified benchmarks~\cite{xia2024sportqa,xia2024sportu,xia2025sportr,he2025finebadminton,rao2025multi,zou2025deepsport} have been proposed to integrate diverse sports-related tasks under a common evaluation framework in the Question-Answer format.
Existing efforts remain largely action-centric, primarily focusing on basic sport rules or high-level semantics in events.
In contrast, our work shifts the focus toward spatial intelligence in sports, emphasizing metrically grounded and human-centric spatial reasoning beyond conventional activity-based evaluation.
\section{CourtSI Dataset}
In this section, we first present the semi-automatic reconstruction data engine that enables scalable dataset construction. We then describe the CourtSI and CourtSI-Bench, which are built upon explicit 3D scene reconstruction.

\subsection{Data Engine}
\label{sec:data-engine}
\begin{figure}[t]
    \centering
    \includegraphics[width=\linewidth]{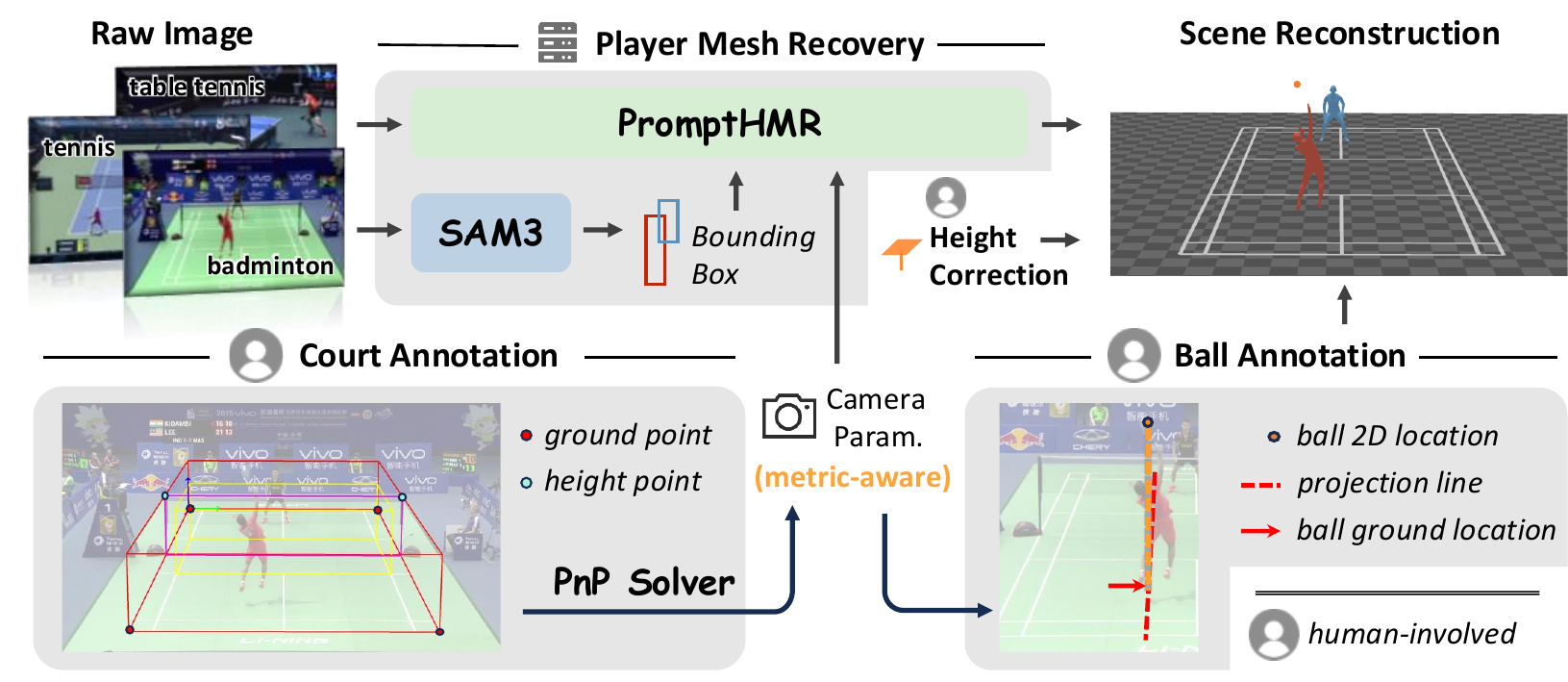}
    \caption{Overview of the data engine. It consists of court annotation for metric-aware camera parameter estimation, ball annotation, and player mesh recovery. By leveraging court geometry and incorporating human-in-the-loop supervision, the system enables accurate and world-grounded reconstruction in sports scenarios.}
    \label{fig:annotation}
\end{figure}

To construct spatial intelligence QA pairs from sports images, we adopt an explicit pipeline that first reconstructs the 3D scene and then formulates questions and derives answers based on the recovered spatial states.
This design enables scalable QA generation, as answers can be computed through deterministic rules grounded in the reconstructed 3D information.
In practice, the primary challenge lies in accurate scene reconstruction, particularly in estimating camera parameters at metric scale and recovering reliable depth for players and balls.

We investigate state-of-the-art monocular methods, including WildCamera~\cite{zhu2023tame} and DepthAnythingV3~\cite{lin2025depth}, but find them insufficiently robust (\cref{sec:supp-compare-mono-methods} for detailed comparisons).
Unlike previous benchmarks~\cite{deng2025internspatial,SpatialRGPT,chen2024spatialvlm,sun2025spacevista}, as illustrated in \cref{fig:annotation}, we develop a human-involved pipeline that exploits court geometry for reliable metric reconstruction. The pipeline consists of the following components:

\paragraph{\textbf{Court Annotation}.}
Sports courts follow standardized geometric layouts, where the real-world dimensions of key structures (e.g., boundary lines and net height) are fixed for each sport. This property allows us to determine the 3D coordinates of predefined court keypoints in a metric world space.
We manually annotate corresponding 2D court keypoints in images, including four ground corner points and two height points on the net.
Given these 2D–3D correspondences, camera parameter calibration naturally becomes a Perspective-n-Point (PnP) problem, in which camera intrinsics and extrinsics are metric-accurately optimized via a PnP solver.
This design defines a unified world coordinate system anchored to the court, while the additional height points on the net stabilize focal length estimation for more reliable reconstruction.
For subsequent spatial intelligence learning, the resulting coordinate system standardizes spatial references across samples, reducing cross-scene variability and enabling consistent localization.

\paragraph{\textbf{Ball Annotation}.}
The ball is typically small, making it difficult for monocular depth estimation models to capture reliably. Moreover, as previously discussed, these models generally lack metric-scale accuracy.
However, as a critical object in sports scenes, precise localization of the ball is essential.
Inspired by~\cite{van20223d}, we design a tool that converts depth estimation into ground projection estimation, which is more intuitive for human annotators.
With known camera parameters, a 2D pixel $\mathbf{p}$ corresponds to a 3D ray in world coordinates, parameterized as:
\begin{equation}
\mathbf{X}(\lambda) = -\mathbf{R}^T \mathbf{t} + \lambda \mathbf{R}^T \mathbf{K}^{-1} \mathbf{p}, \quad \lambda > 0,
\end{equation}
where $\mathbf{K}$ denotes the camera intrinsics, and $\mathbf{R}, \mathbf{t}$ are the extrinsics. $\lambda$ is the depth parameter that varies along the ray.
The projection line with the court plane $Z=0$ intersection is obtained by solving
\begin{equation}
Z(\lambda) = 0.
\end{equation}
Based on this, annotators are instructed to click the 2D position of the ball and its corresponding ground projection along an assistive projection line rendered in the image. Then the depth parameter $\lambda$ of the original ball pixel can be analytically solved, allowing us to recover the 3D location of the ball.

\paragraph{\textbf{Player Mesh Recovery}.}
We adopt the state-of-the-art human mesh recovery method 
Prompt-\\HMR~\cite{phmr} to estimate SMPL-X~\cite{smplx} parameters in the camera coordinate system. The model takes player bounding boxes and camera parameters as input to produce plausible human pose and shape reconstructions. To obtain reliable bounding boxes, we employ SAM3~\cite{carion2025sam} with text prompts and manually refine incorrect detections.
However, we observe that the reconstructed human meshes frequently exhibit inaccurate depth estimation (e.g., foot penetration or floating).
Therefore, we adopt a strategy similar to ball annotation, by annotating the height of the lowest mesh vertex.
The entire mesh is then re-aligned to the correct depth using a perspective transformation based on the annotation.

As introduced above, a sports scene can be reconstructed in a world-grounded manner using our data engine.
Please refer to \cref{sec:supp-pipline-details} for additional details.

\subsection{Dataset Curation}

\paragraph{\textbf{Data Preparation}.}
We build our dataset and benchmark upon broadcast-view images collected from RacketVision~\cite{dong2025racketvision}, a large-scale benchmark containing 1,672 professional net sports clips, including badminton, tennis, and table tennis.
To ensure data quality, we first filter out frames with extreme viewing angles and then apply our data engine to reconstruct 3D scenes from the remaining.

\begin{figure}
    \centering
    \includegraphics[width=\linewidth]{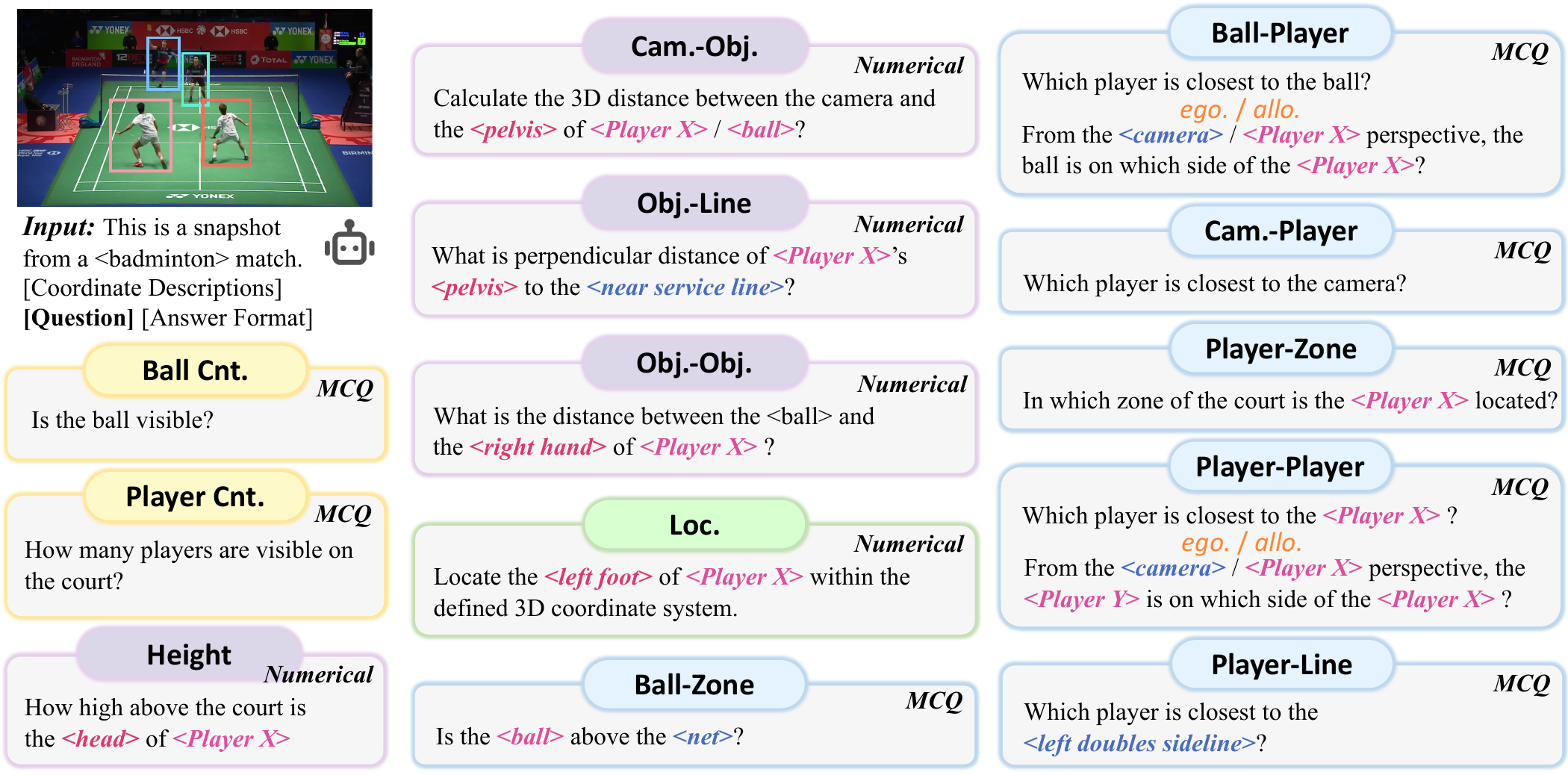}
    \caption{Taxonomy and examples of CourtSI. The questions are categorized into: \colorbox{yellow!30}{spatial counting}, \colorbox{RoyalPurple!20}{distance measurement}, \colorbox{LimeGreen!30}{localization}, and \colorbox{cyan!10}{relational reasoning}.
    Cnt. denotes counting. Obj. refers to object, including the ball and players. Cam. denotes camera. Ego. and Allo. denote to ego-centric and allo-centric views. }
    \label{fig:qa}
\end{figure}

\paragraph{\textbf{Question-Answer Generation}.}
QA pairs are automatically constructed using predefined question templates together with the corresponding 3D reconstruction outputs. 
As illustrated in \cref{fig:qa}, we organize the QA pairs under a unified taxonomy comprising four categories: spatial counting, distance measurement, localization, and relational reasoning.
The questions target core sports entities, including the ball, players, and the court, across camera and world views.
In addition to semantic categorization, the QA pairs cover numerical and multiple-choice questions (MCQs).

To enhance question diversity, we design multiple templates for each question category, resulting in a total of 94 templates. Following \cite{VSI}, each question is accompanied by a general description and an example of the expected answer format to provide clear instructions. Details are provided in \cref{sec:supp-question-template}.

The generated QA pairs exhibit the following characteristics:
(i) \textit{Metric-aware}. Since accurate 3D positions of players and the ball are available, precise metric distance measurement can be performed in real-world units.
(ii) \textit{Human-centric}. Leveraging recovered human meshes, we formulate fine-grained body-part-level questions. Examples include locating a player’s foot or measuring inter-player distance using the pelvis as a reference point, which is commonly treated as the human body center in biomechanics. 
Both egocentric and allocentric perspectives are involved, and all answers are generated automatically based on directional cues from the human mesh.

\begin{figure}[!t]
    \centering
    \includegraphics[width=\linewidth]{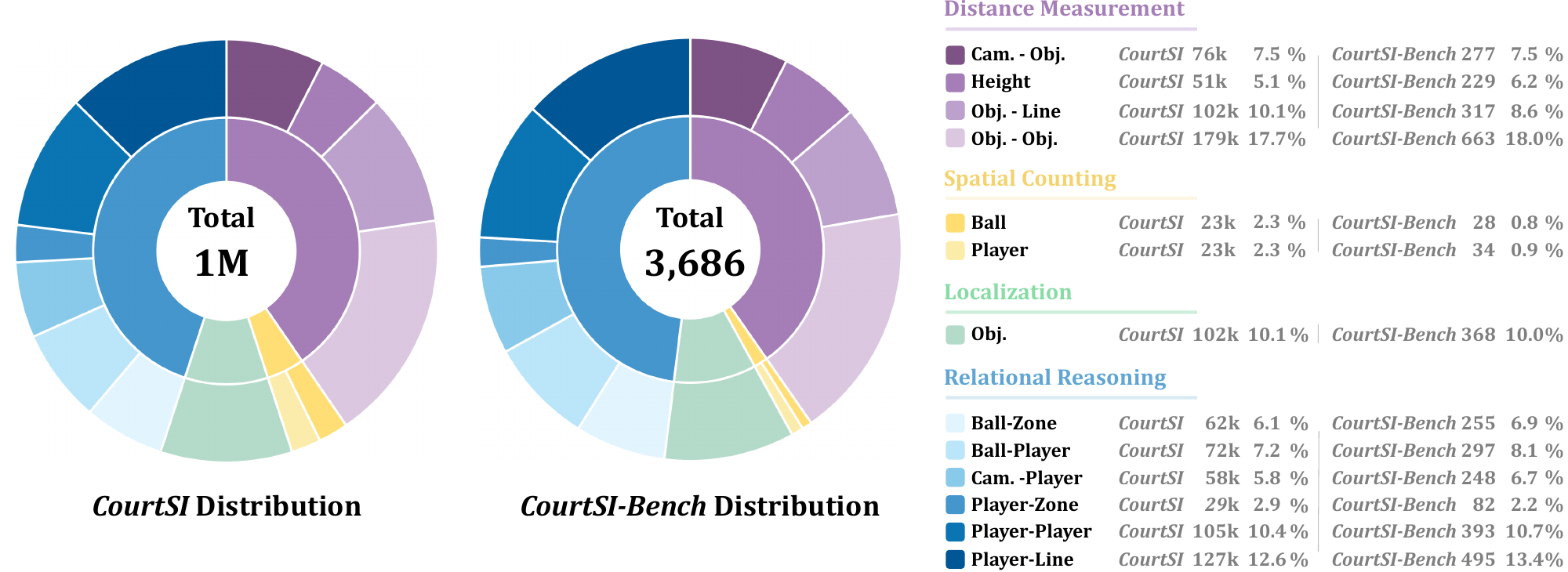}
    \caption{Distribution of CourtSI and CourtSI-Bench. Obj. refers to object, including the ball and players. Cam. denotes camera.}
    \label{fig:distribution}
\end{figure}

We construct CourtSI with 1,008,941 QA pairs generated from 52,481 images spanning 1,057 unique scenes.
In addition, we introduce CourtSI-Bench as a dedicated benchmark, comprising 3,686 QA pairs sampled from 1,988 images across 382 distinct scenes. 
The dataset and benchmark have no scene overlap, preventing potential information leakage.

The distribution of CourtSI and CourtSI-Bench is illustrated in \cref{fig:distribution}.
We carefully balance the categories by considering both their practical importance in sports scenarios and their relative difficulty. 
For CourtSI-Bench, we maintain a relatively balanced distribution of items across different sports to ensure reliable evaluation, as detailed in \cref{sec:supp-data-dist}.

\paragraph{\textbf{Quality Control}.}
To evaluate the reliability of CourtSI and CourtSI-Bench, we first conduct a quality assessment of the data produced by our data engine.
Since ground-truth 3D annotations are unavailable for monocular broadcast videos, we instead leverage a purpose-built multi-view dataset collected by our team, capturing professional matches with camera configurations similar to the source data in RacketVision.
This dataset contains a total of 6,505 frames for each synchronized view.
We use chessboard calibration for camera parameters and apply triangulation to obtain 3D location from annotated 2D ball and player keypoints (details are provided in \cref{sec:supp-compare-mono-methods}).
As shown in table~\ref{tab:quality-control}, the focal length estimation error is approximately 2\%, while both ball and player localization errors remain at the \emph{centimeter level}. 
These results indicate that our data engine produces plausible world-grounded reconstructions. 
Furthermore, the errors are set as a reference for evaluation. For distance measurement, predictions with errors below a predefined threshold are considered correct.

For CourtSI-Bench, we additionally conduct human verification. Two annotators independently review all QA pairs with access to visualizations of the reconstructed scenes. 
This allows them to identify potential reconstruction failures that may lead to incorrect answers. Annotators assess the correctness of each QA pair, and any pair flagged by either annotator is removed.
The process acts as a post-validation for the data engine, ensuring that occasional reconstruction failures do not compromise the overall QA data quality in CourtSI-Bench.

\begin{table}[t]
\centering
\caption{Quantitative error analysis of the data engine. MPJPE denotes Mean Per Joint Position Error for human skeletons.}
\label{tab:quality-control}
\begin{tabular}{c @{\hspace{5pt}} c @{\hspace{4pt}} c @{\hspace{5pt}}c @{\hspace{5pt}}c @{\hspace{4pt}} c @{\hspace{5pt}} c}
\toprule
\multicolumn{2}{c}{\textbf{Camera}} & \multicolumn{3}{c}{\textbf{Ball}} & \multicolumn{2}{c}{\textbf{Player}} \\
\cmidrule(lr){1-2}\cmidrule(lr){3-5}\cmidrule(lr){6-7}
{$f_x$} & {$f_y$} & {$X$} & {$Y$} & {$Z$} & {Pelvis} & {MPJPE} \\
\midrule
$2.2 \%$ & $2.4 \%$ & $22 \mathrm{cm}$ & $9 \mathrm{cm}$ & $9 \mathrm{cm}$ & $23 \mathrm{cm}$ & $17 \mathrm{cm}$ \\
\bottomrule
\end{tabular}
\end{table}
\section{Experiment}
\label{sec:exp}
\begin{table}[!th]
  \centering
  \caption{Quantitative results on CourtSI-Bench. \colorbox{YellowOrange!60}{\textbf{Dark orange}} and \colorbox{YellowOrange!20}{light orange} highlight the best and second-best results within each group of models (proprietary and open-source). \textcolor{teal!90}{---parsed} denotes results obtained by using a LLM to extract answers from the original model outputs.
  Dist. Means., Cnt., Loc., and Rel. denote Distance Measurement, Counting, Localization, and Relational tasks, respectively.}
\label{tab:main_exp}
    \scriptsize
    \begin{tabularx}{\linewidth}
    {%
      l
      *{4}{>{\centering\arraybackslash}p{0.029\linewidth}}  
      *{2}{>{\centering\arraybackslash}p{0.029\linewidth}}  
      *{1}{>{\centering\arraybackslash}p{0.029\linewidth}}  
      *{6}{>{\centering\arraybackslash}p{0.029\linewidth}}  
      *{1}{>{\centering\arraybackslash}p{0.03\linewidth}}  
    }
    \toprule
    \multirow{2}{*}[-5ex]{\scriptsize \textbf{Models}}
      & \multicolumn{4}{c}{\cellcolor{RoyalPurple!15}   \scriptsize \textbf{Dist. Meas.}}
      & \multicolumn{2}{c}{\cellcolor{yellow!30} \scriptsize \textbf{Cnt.}}
      & \cellcolor{LimeGreen!40}  \scriptsize \textbf{Loc.}
      & \multicolumn{6}{c}{\cellcolor{cyan!10}\scriptsize \textbf{Rel. Reasoning}}
      & \multirow{2}{*}[-3ex]{\scriptsize \myrothead{\textbf{Overall}}} \\

    {} & \myrothead{Cam.-Obj.} & \myrothead{Height} & \myrothead{Obj.-Line}
    & \myrothead{Obj.-Obj.}   & \myrothead{Ball}  & \myrothead{Player} & \myrothead{Obj.}
    & \myrothead{Ball-Zone}   & \myrothead{Ball-Player}  & \myrothead{Cam.-Player}
    & \myrothead{Player-Zone}   & \myrothead{Player-Player}  & \myrothead{Player-Line}
    & {} \\
    \midrule
    \rowcolor{gray!20}\multicolumn{15}{l}{\emph{\textbf{Baseline}}} \\
    \addlinespace[0.4em]
    {\textcolor{teal!90}{[5$\%$ Set]} Human} & {64.4} & {92.7} & {67.8} & {70.0} & {100} & {100} & {11.9} & {85.7} & {75.0} & {100} & {83.3} & {90.3} & {88.9} & {73.6} \\
    \midrule
    \rowcolor{gray!20}\multicolumn{15}{l}{\emph{\textbf{Proprietary Models}}} \\
    \addlinespace[0.4em]
    {GPT-5.2} & \cellcolor{YellowOrange!20}{27.9} & {78.4} & {31.8} & \cellcolor{YellowOrange!20}{49.2} & {32.1} & \cellcolor{YellowOrange!20}{100} & \cellcolor{YellowOrange!20}{1.1} & \cellcolor{YellowOrange!20}{68.2} & {67.0} & {75.4} & \cellcolor{YellowOrange!20}{50.0} & {67.7} & \cellcolor{YellowOrange!20}{77.4} & \cellcolor{YellowOrange!20}{53.7} \\
    {Gemini-3-Pro} & {0.0} & {8.7} & {0.0} & {0.0} & {21.4} & {97.1} & {0.0} & 
    {10.6} & {3.0} & {37.5} & {34.1} & {20.4} & {5.3} & {8.7} \\
    {\textcolor{teal!90}{\textit{\hspace*{1em}---parsed}}}& \cellcolor{YellowOrange!60}{\textbf{40.4}} & \cellcolor{YellowOrange!60}{\textbf{81.4}} & \cellcolor{YellowOrange!60}{\textbf{50.5}} & \cellcolor{YellowOrange!60}{\textbf{67.8}} & {60.7} & \cellcolor{YellowOrange!20}{100} & \cellcolor{YellowOrange!60}{\textbf{5.2}} & \cellcolor{YellowOrange!60}{\textbf{71.4}} & \cellcolor{YellowOrange!60}{\textbf{70.7}} & \cellcolor{YellowOrange!60}{\textbf{89.5}} & \cellcolor{YellowOrange!60}{\textbf{73.2}} & \cellcolor{YellowOrange!60}{\textbf{85.0}} & \cellcolor{YellowOrange!60}{\textbf{79.8}} & \cellcolor{YellowOrange!60}{\textbf{64.6}} \\
    {Seed1.8} & {3.0} & {72.7} & {43.8} & {45.2} & \cellcolor{YellowOrange!20}{75.0} & \cellcolor{YellowOrange!20}{100} & {0.5} & {65.5} & \cellcolor{YellowOrange!20}{69.0} & \cellcolor{YellowOrange!20}{82.3} & {40.2} & \cellcolor{YellowOrange!20}{71.0} & \cellcolor{YellowOrange!20}{77.4} & {52.7} \\
    {Claude-Sonnet4.5} & {0.0} & {0.0} & {0.0} & {0.0} & {85.7} & {88.2} & {0.0} & {26.3} & {1.7} & {11.7} & {14.6} & {2.5} & {1.4} & {5.0} \\
    {\textcolor{teal!90}{\textit{\hspace*{1em}---parsed}}}& {19.4} & \cellcolor{YellowOrange!20}{80.7} & \cellcolor{YellowOrange!20}{44.5} & \cellcolor{YellowOrange!20}{49.2} & \cellcolor{YellowOrange!60}{\textbf{85.7}} & {97.1} & {0.3} & {58.4} & {55.2} & {61.3} & {47.6} & {61.3} & {60.2} & {49.1} \\
    {Grok4} & {12.2} & {60.0} & {30.9} & {36.7} & {50.0} & {97.1} & {0.0} & {44.7} & {38.7} & {37.5} & {34.1} & {46.8} & {48.9} & {36.2} \\
    {Qwen3-Max} & {9.5} & {72.4} & {23.4} & {35.0} & {7.1} & {91.2} & {0.0} & {52.9} & {48.1} & {55.2} & {13.4} & {50.4} & {51.3} & {38.2} \\
    \midrule
    \rowcolor{gray!20}\multicolumn{15}{l}{\emph{\textbf{Open-source General Models}}} \\
    \addlinespace[0.4em]
    {Qwen3-VL-8B} & {3.1} & {49.3} & {21.3} & {27.1} & {39.3} & {97.1} & {0.0} & {56.9} & {57.9} & {71.8} & {30.5} & {52.9} & {50.1} & {37.7} \\
    {Qwen3-VL-32B} & {4.1} & \cellcolor{YellowOrange!20}{60.7} & {5.1} & {22.6} & {39.3} & \cellcolor{YellowOrange!20}{100} & {0.0} & \cellcolor{YellowOrange!20}{64.7} & {56.6} & {76.2} & {48.8} & {57.8} & {64.2} & {39.8} \\
    {Qwen3-VL-235B-A22B} & {1.2} & {58.9} & \cellcolor{YellowOrange!20}{24.3} & \cellcolor{YellowOrange!60}{\textbf{34.9}} & {42.9} & \cellcolor{YellowOrange!20}{100} & {0.0} & \cellcolor{YellowOrange!60}{\textbf{67.5}} & \cellcolor{YellowOrange!60}{\textbf{70.0}} & \cellcolor{YellowOrange!60}{\textbf{84.3}} & {35.3} & \cellcolor{YellowOrange!60}{\textbf{71.0}} & \cellcolor{YellowOrange!60}{\textbf{71.1}} & \cellcolor{YellowOrange!60}{\textbf{47.2}} \\
    {InternVL3.5-8B} & {0.0} & {0.0} & {0.0} & {0.0} & \cellcolor{YellowOrange!60}{\textbf{78.6}} & {67.6} & {0.0} & {50.2} & {55.6} & {69.8} & {20.7} & {51.4} & {60.0} & {27.9} \\
    {InternVL3.5-38B} & {0.0} & {0.0} & {0.0} & {0.5} & {42.9} & \cellcolor{YellowOrange!20}{100} & {0.0} & {58.4} & {64.6} & {79.8} & {32.9} & {63.1} & \cellcolor{YellowOrange!20}{67.7} & {32.5} \\
    {InternVL3.5-241B-A28B} & {0.7} & {51.9} & {16.5} & {16.0} & {39.3} & \cellcolor{YellowOrange!20}{100} & {0.0} & {58.4} & \cellcolor{YellowOrange!20}{66.0} & \cellcolor{YellowOrange!20}{80.2} & \cellcolor{YellowOrange!20}{56.1} & {64.1} & {65.3} & {40.0} \\
    {Kimi-VL-16B-A3B} & {0.0} & {56.4} & {19.5} & {16.7} & {46.4} & \cellcolor{YellowOrange!20}{100} & {0.0} & {56.5} & {57.6} & {60.5} & {32.9} & {51.1} & {47.9} & {34.7} \\
    {LLaVA-OneVision-7B} & {0.0} & {45.2} & {14.9} & {16.0} & {46.4} & \cellcolor{YellowOrange!20}{100} & {0.0} & {56.0} & {50.5} & {73.4} & {41.5} & {53.2} & {51.5} & {34.6} \\
    {LLaVA-OneVision-72B} & \cellcolor{YellowOrange!60}{\textbf{13.5}} & \cellcolor{YellowOrange!60}{\textbf{67.1}} & \cellcolor{YellowOrange!60}{\textbf{24.7}} & \cellcolor{YellowOrange!20}{29.5} & {28.6} & \cellcolor{YellowOrange!20}{100} & \cellcolor{YellowOrange!20}{0.3} & {54.9} & {61.6} & {72.2} & {54.9} & {55.7} & {55.6} & \cellcolor{YellowOrange!20}{42.0} \\
    {LLaVA-OneVision1.5-8B} & {3.7} & {49.0} & {21.2} & {26.9} & {10.7} & \cellcolor{YellowOrange!20}{100} & \cellcolor{YellowOrange!20}{0.3} & {44.7} & {44.1} & {56.0} & {34.1} & {45.3} & {46.9} & {33.3} \\
    \midrule
    \rowcolor{gray!20}\multicolumn{15}{l}{\emph{\textbf{Open-source Spatial Intelligence Models}}} \\
    \addlinespace[0.4em]
    {\textcolor{teal!90}{[Base]} Qwen2.5-VL-7B} & \cellcolor{YellowOrange!20}{4.8} & {50.5} & {20.2} & {9.3} & {35.7} & \cellcolor{YellowOrange!20}{100} & {0.0} & {54.9} & {60.3} & {74.2} & \cellcolor{YellowOrange!60}{\textbf{58.5}} & {61.6} & {54.9} & {37.0} \\
    {SpaceR-7B} & {0.4} & {47.5} & {3.9} & {1.9} & {39.2} & \cellcolor{YellowOrange!20}{100} & {0.0} & {59.6} & {58.6} & {72.2} & {40.2} & {59.2} & {52.3} & {32.8} \\
    {VST-7B-SFT} & {0.0} & {55.2} & {19.3} & {19.7} & {35.7} & \cellcolor{YellowOrange!20}{100} & {0.0} & {51.8} & {57.6} & {78.6} & {48.8} & \cellcolor{YellowOrange!20}{65.9} & {61.0} & {39.6} \\
    {VST-7B-RL} & {0.0} & {50.3} & {22.2} & {20.3} & {35.7} & \cellcolor{YellowOrange!20}{100} & {0.0} & {54.9} & {59.6} & {75.4} & {53.6} & {64.9} & {61.8} & {40.0} \\
    \arrayrulecolor{gray!60}\midrule
    \arrayrulecolor{black}
    {\textcolor{teal!90}{[Base]} Qwen2.5-VL-3B} & {4.6} & {51.4} & {20.3} & {20.1} & {35.7} & {97.1} & {0.0} & {52.9} & {49.8} & {68.5} & {40.2} & {51.7} & {46.6} & {35.0} \\
    {SpatialLadder} & {0.0} & {56.7} & {22.3} & {12.4} & {57.1} & {97.1} & {0.0} & {53.7} & {50.5} & {63.7} & {48.8} & {55.7} & {49.5} & {34.7} \\
    \arrayrulecolor{gray!60}\midrule
    \arrayrulecolor{black}
    {\textcolor{teal!90}{[Base]} InternVL3-8B} & {0.0} & {0.0} & {0.0} & {0.0} & {46.4} & {14.7} & {0.0} & {57.3} & {57.9} & {71.0} & {31.7} & {52.4} & {56.6} & {27.8} \\
    {SenseNova-SI-8B} & {0.7} & {40.0} & {21.3} & {17.5} & \cellcolor{YellowOrange!20}{67.9} & {47.1} & {0.0} & {43.5} & {53.5} & {49.2} & {26.8} & {49.4} & {48.9} & {31.5} \\
    \arrayrulecolor{gray!60}\midrule
    \arrayrulecolor{black}
    {Cambrain-S-7B} & {0.0} & {3.2} & {0.2} & {0.0} & {17.9} & {85.3} & {0.0} & {63.5} & {44.8} & {58.1} & {7.3} & {55.2} & {47.5} & {25.5} \\
    \midrule
    \textbf{Ours}$_{\ \textbf{\text{Qwen3-VL-8B}}}$ & {60.2} & {94.2} & {47.6} & {68.4} & {92.9} & {100} &
    {7.9} & {65.1} & {63.6} & {78.2} & {85.4} & {56.7} & {68.5} & {61.2}  \\
    {\emph{Improvement}} & {\emph{57.1}} & {\emph{44.9}} & {\emph{26.3}} & {\emph{41.3}} & {\emph{53.6}} & {\emph{2.9}} & {\emph{7.9}} & {\emph{8.2}} & {\emph{5.7}} & {\emph{6.4}} & {\emph{54.9}} & {\emph{3.8}} & {\emph{18.4}} & {\emph{23.5}}\\
    \bottomrule
    \end{tabularx}
\end{table}

\subsection{Evaluation Setup}
\paragraph{\textbf{Baseline Models}.}
We conduct a comprehensive evaluation of 25 state-of-the-art vision-language models (VLMs), spanning diverse model families and parameter scales.
For proprietary models, we include GPT-5.2, Gemini-3-Pro, Seed1.8, Claude-Sonnet4.5, Grok4, and Qwen3-Max.
For open-source models, we evaluate the Qwen3-VL series~\cite{qwen3vl}, InternVL3.5 series~\cite{wang2025internvl35}, Kimi-VL~\cite{team2025kimi}, and the LLaVA-OneVision series~\cite{li2024llava,an2025llava}. 
In addition, we benchmark models fine-tuned on prior spatial intelligence datasets, including SpaceR~\cite{ouyang2025spacer}, VST~\cite{yang2025visual}, SpatialLadder~\cite{li2026spatialladder}, SenseNova-SI~\cite{Sensenova}, and Cambrain-S~\cite{yang2025cambrian}, together with their corresponding base models~\cite{qwen25vl,zhu2025internvl3}.
Human performance is reported as a reference for the benchmark.
Finally, to assess the task-specific learning potential of CourtSI, we further conduct supervised fine-tuning (SFT) on Qwen3-VL-8B. 
The model is trained for one epoch using a global batch size of 2048 and a learning rate of $5\times10^{-6}$ in LLaMA Factory environment~\cite{zheng2024llamafactory}.
Please refer to \cref{sec:supp-exp} for more details.

\paragraph{\textbf{Evaluation Metrics.}}
Following VSI~\cite{VSI}, we use \emph{Accuracy} based on exact matching as the main metric. For numerical answer tasks in distance measurement and localization, we report Threshold Mean Relative Accuracy(T-MRA) to allow for a certain error:

\newpage

\begin{equation}
    \text{T-MRA} = \frac{1}{10}\sum_{\theta \in \mathcal{C}} 
\mathbbm{1} \left( \frac{|\hat{y} - y| - T}{y} < 1 - \theta \right),
\end{equation}
where $y$ and $\hat{y}$ denote ground truth and prediction, respectively. The confidence thresholds span $\{0.5, 0.55, ..., 0.95\}$, consistent with VSI. The distance threshold $T$ is set to $15 \mathrm{cm}$ according to \cref{tab:quality-control}.

\subsection{Evaluation on CourtSI-Bench}
We evaluate baseline models on CourtSI-Bench. Each input consists of a question paired with a single image annotated with bounding boxes and corresponding instructions to differentiate among players~\cite{deng2025internspatial}.
The results are summarized in \cref{tab:main_exp}.
We provide a detailed analysis below.

\paragraph{\textbf{Human Level Performance}.}
We recruit two volunteers to complete the evaluation on a uniformly sampled $5\%$ subset of CourtSI-Bench.
Human evaluators achieve the strongest performance compared to all existing models across all metrics. 
However, even with court geometry as a reference, human performance drops noticeably on metric-sensitive tasks, particularly distance measurement and localization.
This limitation is also observed in several 3D vision tasks, where humans are required to estimate absolute distances and tend to underperform state-of-the-art specialized models.
The current state of spatial intelligence in sports scenarios motivates the development of more general models capable of accurate 3D perception and reasoning under flexible language instructions, thereby assisting humans in metric-level spatial understanding.

\paragraph{\textbf{Proprietary Models.}}
Several proprietary models demonstrate strong performance, in some cases approaching human-level results. 
Among them, Gemini-Pro achieves the best overall performance across most metrics, with the exception of ball counting.
However, we observe notable issues with instruction compliance in Gemini3-Pro and Claude-Sonnet-4.5. Although the models are required to produce final answers in a specified format, they frequently generate uncontrolled intermediate reasoning or extended explanations, violating the output constraints.
Notably, their competitive performance is largely achieved only after applying an additional LLM to parse answers from the original outputs. Without this post-processing step, the performance drops significantly, indicating substantial room for improvement in controllable response generation.

\paragraph{\textbf{Open-source General Models}.}
Among the open-source general models, Qwen3-VL-235B-A22B achieves the strongest performance, with only a limited gap compared to the best-performing proprietary models. However, most open-source models perform poorly on CourtSI-Bench, with overall accuracy below 40\%.
Moreover, in distance measurement tasks, some models even exhibit near-total failure under the loose T-MRA metric.

\paragraph{\textbf{Open-source Spatial Intelligence Models}.}
For spatial intelligence models, although they are specifically fine-tuned for spatial relationship understanding and metric distance measurement, we do not observe consistent improvements over their respective base models on CourtSI-Bench.
This suggests that sports scenarios introduce additional spatial reasoning challenges that are not sufficiently captured by existing large-scale spatial intelligence benchmarks.

\paragraph{\textbf{SFT on CourtSI}.}
After conducting SFT on CourtSI, the Qwen3-VL-8B model gains consistent improvement across all evaluation metrics, achieving a gain of $23.5$ percentage points in accuracy. Notably, performance on the challenging distance measurement task improves by more than 25 percentage points. These results demonstrate the effectiveness of CourtSI in enhancing the spatial intelligence of VLM in sports.

\subsection{In-depth Error Analysis on CourtSI-Bench}
\begin{figure}[t]
    \centering
    \includegraphics[width=\linewidth]{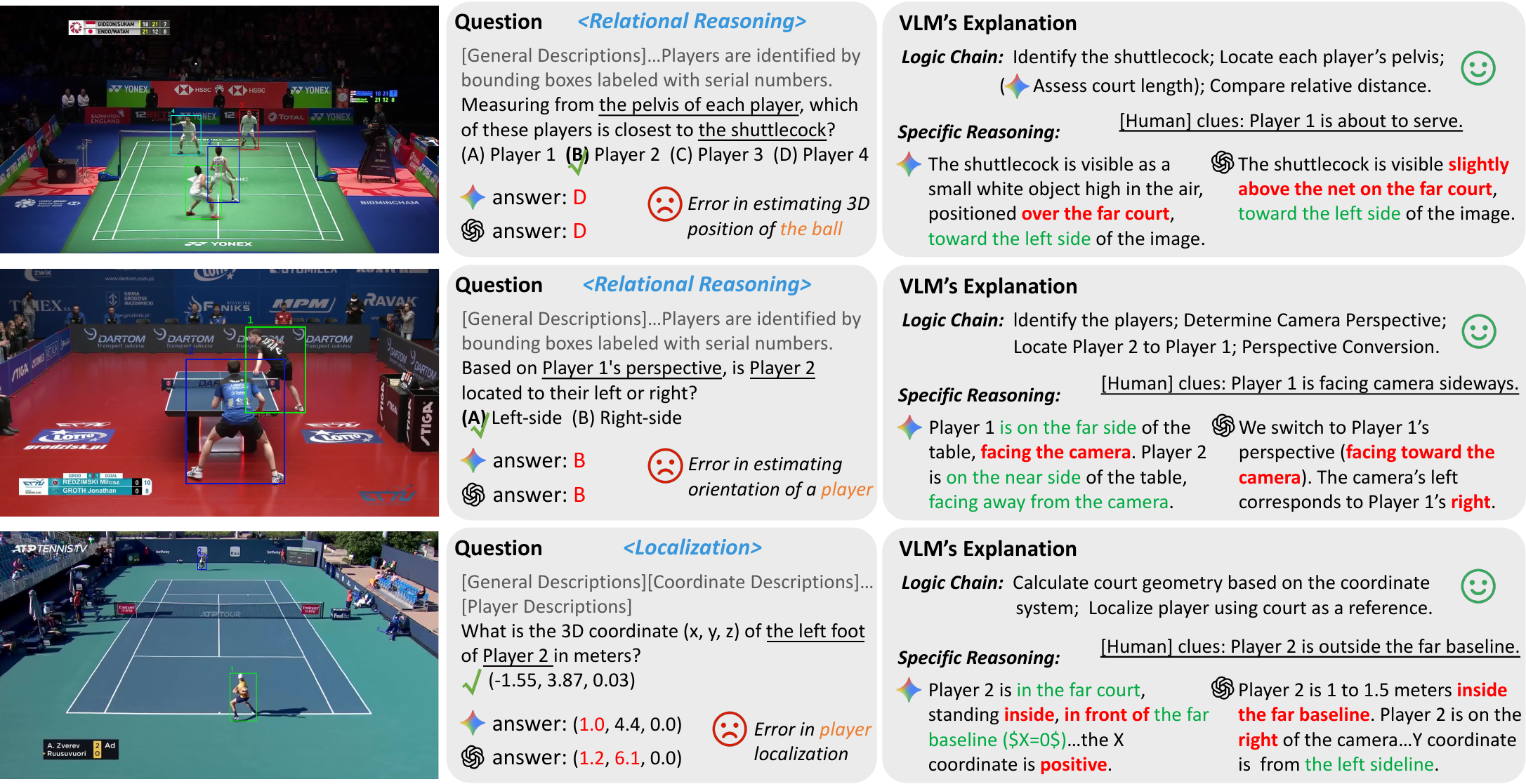}
    \caption{Error Analysis. The VLMs are prompted to provide detailed step-by-step reasoning. Correct and incorrect reasoning steps are highlighted in \textcolor{ForestGreen}{green} and \textcolor{red}{red}, respectively. Questions and VLM's explanations are simplified for demonstration.}
    \label{fig:error-analysis}
\end{figure}

To better understand VLM performance on CourtSI-Bench, we conduct case studies on the categories with low accuracy, including relational reasoning and localization. 
Specifically, we prompt the strongest-performing Gemini-3-Pro and GPT-5.2 to explain the reasoning behind their predictions on failure cases.

We summarize the representative cases in \cref{fig:error-analysis}. From top to bottom, the cases involve: the relative distance between sport-specific objects, ball and player; reasoning about player–player relationships under an allo-centric perspective; metric-aware localization for absolute distance measurement.

In many instances, the VLMs produce human-like and logically structured reasoning chains. For example, they first localize relevant objects before comparing their spatial relationships.
Furthermore, VLMs perform well under some challenging instructions: they identify the tiny ball (``\textit{small white object}'', \cref{fig:error-analysis}, top), handle ego-centric to allo-centric perspective conversion (``\textit{switch to ...from camera}``, \cref{fig:error-analysis}, mid), and leverage court geometry as a reference based on general sport knowledge (``\textit{the far baseline (X=0)}'', \cref{fig:error-analysis}, bottom).
These observations suggest that VLMs can interpret spatial descriptions from instructions and demonstrate a basic level of structured reasoning.

However, the models struggle with accurate 3D localization from 2D imagery and fine-grained relational understanding. 
In the top and bottom cases of \cref{fig:error-analysis}, the VLMs incorrectly estimate object relationships with respect to the court geometry. 
In the middle case, a counterfactual configuration, where a player stands on the far side of the court while facing sideways relative to the camera, leads to erroneous results.
These failure modes stem from the distinctive characteristics of the curated CourtSI-Bench, which introduce spatial ambiguities that challenge current VLMs and highlight substantial room for improvement.

\begin{wrapfigure}{r}{0.52\linewidth}
    \centering
    \vspace{-20pt}
    \includegraphics[width=\linewidth]{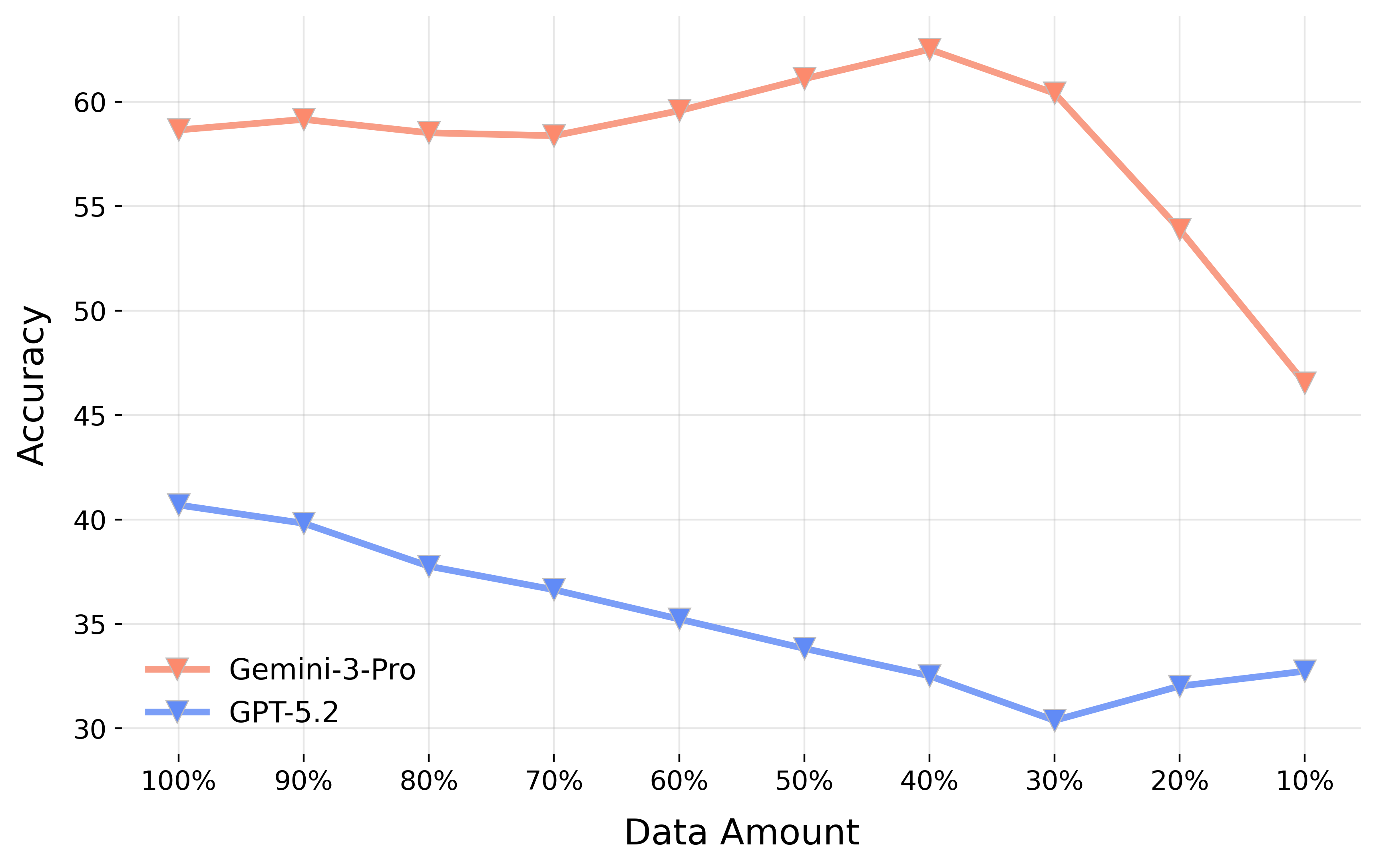}
    \caption{The impact of perspective ambiguity.}
    \vspace{-10pt}
    \label{fig:adv-curve}
\end{wrapfigure}

In addition, we further demonstrate VLMs’ limitations in handling spatial ambiguity caused by perspective projection.
Specifically, we rank the cases of distance measurement in CourtSI-Bench by a ratio of 3D distance to 2D distance, to reflect the level of perspective ambiguity.
A higher ratio indicates that objects are distant in 3D space but appear close in the image plane due to perspective effects.
As shown in \cref{fig:adv-curve}, we evaluate VLMs' performance on the top-percentage subsets. The results show a clear performance degradation as the ambiguity increases, revealing a factor behind the erroneous relational reasoning, particularly for distance measurement when precise estimates are required.

\subsection{Expanding the Scope of CourtSI-Bench}
To broaden the evaluation scope of CourtSI-Bench, we extend it toward more application-oriented settings.
Specifically, we construct two additional scenarios: (i) an unseen-sport evaluation set, CourtSI-Ext, designed to assess the generalization capability of spatial intelligence models fine-tuned on CourtSI data; and (ii) a spatial-aware commentary generation task, which serves as a potential downstream application of models equipped with sports spatial reasoning ability.

\begin{table}[!t]
  \centering

  \caption{Evaluation on CourtSI-Ext. Annotation conventions follow \cref{tab:main_exp}.}
    \label{tab:extend_exp}
    \scriptsize
    \begin{tabularx}{\linewidth}
    {%
      l
      *{4}{>{\centering\arraybackslash}p{0.029\linewidth}}  
      *{2}{>{\centering\arraybackslash}p{0.029\linewidth}}  
      *{1}{>{\centering\arraybackslash}p{0.029\linewidth}}  
      *{6}{>{\centering\arraybackslash}p{0.029\linewidth}}  
      *{1}{>{\centering\arraybackslash}p{0.03\linewidth}}  
    }
    \toprule
    \multirow{2}{*}[-5ex]{\scriptsize \textbf{Models}}
      & \multicolumn{4}{c}{\cellcolor{RoyalPurple!15}   \scriptsize \textbf{Dist. Meas.}}
      & \multicolumn{2}{c}{\cellcolor{yellow!30} \scriptsize \textbf{Cnt.}}
      & \cellcolor{LimeGreen!40}  \scriptsize \textbf{Loc.}
      & \multicolumn{6}{c}{\cellcolor{cyan!10}\scriptsize \textbf{Rel. Reasoning}}
      & \multirow{2}{*}[-3ex]{\scriptsize \myrothead{\textbf{Overall}}} \\

    {} & \myrothead{Cam.-Obj.} & \myrothead{Height} & \myrothead{Obj.-Line}
    & \myrothead{Obj.-Obj.}   & \myrothead{Ball}  & \myrothead{Player} & \myrothead{Obj.}
    & \myrothead{Ball-Zone}   & \myrothead{Ball-Player}  & \myrothead{Cam.-Player}
    & \myrothead{Player-Zone}   & \myrothead{Player-Player}  & \myrothead{Player-Line}
    & {} \\
    \midrule
    {GPT-5.2} & \cellcolor{YellowOrange!20}{61.8} & \cellcolor{YellowOrange!20}{70.0} & \cellcolor{YellowOrange!20}{44.2} & \cellcolor{YellowOrange!20}{53.6} & {33.3} & {100} & {0.0} & \cellcolor{YellowOrange!20}{72.7} & {66.7} & \cellcolor{YellowOrange!20}{90.0} & {21.4} & \cellcolor{YellowOrange!60}{\textbf{73.5}} & \cellcolor{YellowOrange!20}{76.2} & \cellcolor{YellowOrange!20}{55.0} \\
    {Gemini-3-Pro} & {0.0} & {15.4} & {0.0} & {0.0} & {33.3} & {100} & {0.0} & 
    {9.1} & {11.1} & {30.0} & {64.3} & {17.6} & {0.0} & {13.5} \\
    {\textcolor{teal!90}{\textit{\hspace*{1em}---parsed}}}& \cellcolor{YellowOrange!60}{\textbf{75.5}} & \cellcolor{YellowOrange!60}{\textbf{83.1}} & \cellcolor{YellowOrange!60}{\textbf{66.3}} & \cellcolor{YellowOrange!60}{\textbf{56.4}} & \cellcolor{YellowOrange!60}{\textbf{83.3}} & {100} & {0.0} & \cellcolor{YellowOrange!60}{\textbf{90.9}} & \cellcolor{YellowOrange!60}{\textbf{100}} & \cellcolor{YellowOrange!20}{90.0} & \cellcolor{YellowOrange!60}{\textbf{85.7}} & \cellcolor{YellowOrange!20}{70.6} & \cellcolor{YellowOrange!20}{76.2} & \cellcolor{YellowOrange!60}{\textbf{66.8}}\\
    {Qwen3-VL-235B-A22B\ } & {0.0} & {50.8} & {30.4} & {27.6} & \cellcolor{YellowOrange!20}{50.0} & {100} & {0.0} & {63.6} & \cellcolor{YellowOrange!20}{88.9} & \cellcolor{YellowOrange!60}{\textbf{100}} & {64.3} & {67.6} & {71.4} & {47.9} \\
    {LLaVA-OneVision-72B\ } & {19.1} & {53.8} & {28.3} & {21.8} & \cellcolor{YellowOrange!20}{50.0} & {100} & {0.0} & {45.5} & {55.6} & \cellcolor{YellowOrange!20}{90.0} & \cellcolor{YellowOrange!20}{71.4} & {61.8} & {57.1} & {43.3}\\
    \midrule
    {\textcolor{teal!90}{[Base]} Qwen3-VL-8B} & {0.9} & {50.0} & {31.3} & {15.2} & {33.3} & {100} & {0.0} & {63.6} & {77.8} & {100} & {21.4} & {52.9} & {52.4} & {38.2} \\ 
    \textbf{Ours} & {70.0} & {83.1} & {34.6} & {62.7} & {83.3} & {100} & {0.0} & {63.6} & {66.7} & {70.0} & {28.6} & {44.1} & {66.7} & {51.4} \\
    \bottomrule
    \end{tabularx}
\end{table}
\paragraph{\textbf{CourtSI-Ext}.}
Following the taxonomy of CourtSI-Bench, we leverage the data engine in~\cref{sec:data-engine} to construct an extended evaluation set, CourtSI-Ext. It is built on pickleball, a net sport with court geometry similar to tennis and badminton.
CourtSI-Ext contains 215 QA pairs from 111 images across 35 distinct scenes for cross-sport evaluation. We report results of top-performing VLMs from CourtSI-Bench on this extension.
The evaluation process is consistent with CourtSI-Bench. Image examples are presented in \cref{sec:supp-courtsi-ext}.

As shown in \cref{tab:extend_exp}, our fine-tuned model achieves $13.2$ percentage points improvements in overall accuracy compared to its base model, further validating the effectiveness of the curated CourtSI data.
However, the cross-sport generalization challenge in spatial intelligence remains. The improvement of SFT shrinks on CourtSI-Ext. Specifically, in the localization task, although our model reduces the average error to $3.9$ meters compared to about $6$ meters from other baselines, it does not yield corresponding gains in accuracy.
As an initial study, we highlight the cross-sport challenge and curate CourtSI-Ext to serve as a small yet valuable benchmark for broader community validation.

\begin{figure}[t]
    \centering
    \includegraphics[width=\linewidth]{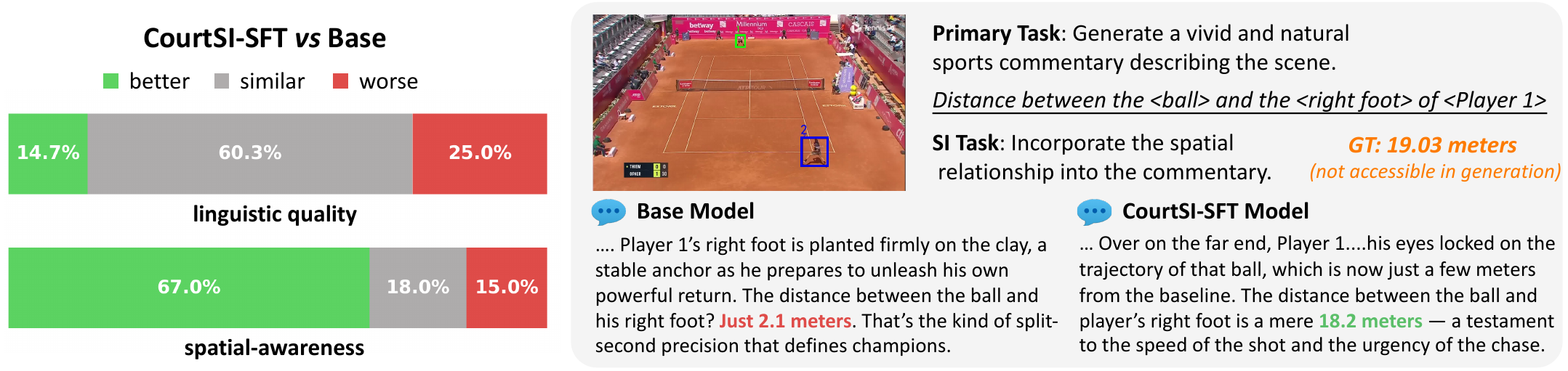}
    \caption{Evaluation on the spatial-aware commentary generation. Comparison between Qwen3-VL-8B fine-tuned on CourtSI and its base model. The left panel shows user study results assessing the quality of generated commentaries in both linguistic and spatial-awareness dimensions. The right panel presents an illustrative example.}
    \label{fig:commentary}
\end{figure}

\paragraph{\textbf{Spatial-aware Commentary Generation}.}
As shown in \cref{fig:commentary} (right), we extract spatial relationships from CourtSI-Bench and instruct models to generate sports commentary that incorporates these spatial relationships. We compare the Qwen3-VL-8B model fine-tuned on CourtSI with its base model. A total of 100 generated commentaries are evaluated through a user study involving three volunteers across both linguistic quality and spatial awareness dimensions.

The results show that fine-tuning on CourtSI significantly improves spatial awareness, while preserving overall linguistic quality, highlighting that the model is able to transfer the spatial capability to downstream commentary generation tasks.
It illustrates the potential of CourtSI to enhance spatial reasoning in sports understanding and to serve as supervision for general VLM post-training.
\section{Conclusion}

In this paper, we present CourtSI, the first large-scale spatial intelligence dataset for sports, comprising over 1M QA pairs, along with the high-quality CourtSI-Bench for evaluation.
By leveraging court geometry as metric anchors, we develop a semi-automatic data engine to produce accurate and scalable supporting data.
A comprehensive evaluation across 25 state-of-the-art VLMs reveals a clear human–AI performance gap and limited generalization from existing spatial intelligence benchmarks. Furthermore, through fine-tuning, cross-sport evaluation, and commentary generation, we broaden the evaluation scope of CourtSI-Bench and demonstrate that CourtSI serves as an effective pathway for advancing the spatial intelligence of VLMs in sports.

\newpage

{
\small
\bibliographystyle{unsrt}
\bibliography{ref}

@inproceedings{VSI,
  title={Thinking in space: How multimodal large language models see, remember, and recall spaces},
  author={Yang, Jihan and Yang, Shusheng and Gupta, Anjali W and Han, Rilyn and Fei-Fei, Li and Xie, Saining},
  booktitle={Proceedings of the Computer Vision and Pattern Recognition Conference},
  pages={10632--10643},
  year={2025}
}

@article{MMSI,
  title={Mmsi-bench: A benchmark for multi-image spatial intelligence},
  author={Yang, Sihan and Xu, Runsen and Xie, Yiman and Yang, Sizhe and Li, Mo and Lin, Jingli and Zhu, Chenming and Chen, Xiaochen and Duan, Haodong and Yue, Xiangyu and others},
  journal={arXiv preprint arXiv:2505.23764},
  year={2025}
}

@article{SPAR,
  title={From flatland to space: Teaching vision-language models to perceive and reason in 3d},
  author={Zhang, Jiahui and Chen, Yurui and Zhou, Yanpeng and Xu, Yueming and Huang, Ze and Mei, Jilin and Chen, Junhui and Yuan, Yu-Jie and Cai, Xinyue and Huang, Guowei and others},
  journal={arXiv preprint arXiv:2503.22976},
  year={2025}
}

@article{SpatialRGPT,
  title={Spatialrgpt: Grounded spatial reasoning in vision-language models},
  author={Cheng, An-Chieh and Yin, Hongxu and Fu, Yang and Guo, Qiushan and Yang, Ruihan and Kautz, Jan and Wang, Xiaolong and Liu, Sifei},
  journal={Advances in Neural Information Processing Systems},
  volume={37},
  pages={135062--135093},
  year={2024}
}

@article{Sensenova,
  title={Scaling spatial intelligence with multimodal foundation models},
  author={Cai, Zhongang and Wang, Ruisi and Gu, Chenyang and Pu, Fanyi and Xu, Junxiang and Wang, Yubo and Yin, Wanqi and Yang, Zhitao and Wei, Chen and Sun, Qingping and others},
  journal={arXiv preprint arXiv:2511.13719},
  year={2025}
}

@inproceedings{MindCube,
  title={Spatial mental modeling from limited views},
  author={Yin, Baiqiao and Wang, Qineng and Zhang, Pingyue and Zhang, Jianshu and Wang, Kangrui and Wang, Zihan and Zhang, Jieyu and Chandrasegaran, Keshigeyan and Liu, Han and Krishna, Ranjay and others},
  booktitle={Structural Priors for Vision Workshop at ICCV'25},
  year={2025}
}

@inproceedings{phmr,
  title={Prompthmr: Promptable human mesh recovery},
  author={Wang, Yufu and Sun, Yu and Patel, Priyanka and Daniilidis, Kostas and Black, Michael J and Kocabas, Muhammed},
  booktitle={Proceedings of the computer vision and pattern recognition conference},
  pages={1148--1159},
  year={2025}
}

@inproceedings{smplx,
  title={Expressive body capture: 3d hands, face, and body from a single image},
  author={Pavlakos, Georgios and Choutas, Vasileios and Ghorbani, Nima and Bolkart, Timo and Osman, Ahmed AA and Tzionas, Dimitrios and Black, Michael J},
  booktitle={Proceedings of the IEEE/CVF conference on computer vision and pattern recognition},
  pages={10975--10985},
  year={2019}
}

@article{xia2025sportr,
  title={SportR: A Benchmark for Multimodal Large Language Model Reasoning in Sports},
  author={Xia, Haotian and Ge, Haonan and Zou, Junbo and Choi, Hyun Woo and Zhang, Xuebin and Suradja, Danny and Rui, Botao and Tran, Ethan and Jin, Wendy and Ye, Zhen and others},
  journal={arXiv preprint arXiv:2511.06499},
  year={2025}
}

@article{xia2024sportu,
  title={Sportu: A comprehensive sports understanding benchmark for multimodal large language models},
  author={Xia, Haotian and Yang, Zhengbang and Zou, Junbo and Tracy, Rhys and Wang, Yuqing and Lu, Chi and Lai, Christopher and He, Yanjun and Shao, Xun and Xie, Zhuoqing and others},
  journal={arXiv preprint arXiv:2410.08474},
  year={2024}
}

@inproceedings{rao2025towards,
  title={Towards universal soccer video understanding},
  author={Rao, Jiayuan and Wu, Haoning and Jiang, Hao and Zhang, Ya and Wang, Yanfeng and Xie, Weidi},
  booktitle={Proceedings of the Computer Vision and Pattern Recognition Conference},
  pages={8384--8394},
  year={2025}
}

@inproceedings{gao2025fsbench,
  title={Fsbench: A figure skating benchmark for advancing artistic sports understanding},
  author={Gao, Rong and Liu, Xin and Hu, Zhuozhao and Xing, Bohao and Xia, Baiqiang and Yu, Zitong and K{\"a}lvi{\"a}inen, Heikki},
  booktitle={Proceedings of the Computer Vision and Pattern Recognition Conference},
  pages={13595--13605},
  year={2025}
}

@inproceedings{he2025finebadminton,
  title={Finebadminton: A multi-level dataset for fine-grained badminton video understanding},
  author={He, Xusheng and Liu, Wei and Ma, Shanshan and Liu, Qian and Ma, Chenghao and Wu, Jianlong},
  booktitle={Proceedings of the 33rd ACM International Conference on Multimedia},
  pages={12776--12783},
  year={2025}
}

@inproceedings{dong2025racketvision,
  title={RacketVision: A Multiple Racket Sports Benchmark for Unified Ball and Racket Analysis},
  author={Dong, Linfeng and Yang, Yuchen and Wu, Hao and Wang, Wei and Hou, Yuenan and Zhong, Zhihang and Sun, Xiao},
  booktitle={Proceedings of the AAAI Conference on Artificial Intelligence},
  year={2026}
}

@article{qwen3vl,
  title={Qwen3-vl technical report},
  author={Bai, Shuai and Cai, Yuxuan and Chen, Ruizhe and Chen, Keqin and Chen, Xionghui and Cheng, Zesen and Deng, Lianghao and Ding, Wei and Gao, Chang and Ge, Chunjiang and others},
  journal={arXiv preprint arXiv:2511.21631},
  year={2025}
}

@article{qwen25vl,
      title={Qwen2.5-VL Technical Report}, 
      author={Shuai Bai and Keqin Chen and Xuejing Liu and Jialin Wang and Wenbin Ge and Sibo Song and Kai Dang and Peng Wang and Shijie Wang and Jun Tang and Humen Zhong and Yuanzhi Zhu and Mingkun Yang and Zhaohai Li and Jianqiang Wan and Pengfei Wang and Wei Ding and Zheren Fu and Yiheng Xu and Jiabo Ye and Xi Zhang and Tianbao Xie and Zesen Cheng and Hang Zhang and Zhibo Yang and Haiyang Xu and Junyang Lin},
      journal={arXiv preprint arXiv:2502.13923},
      year={2025}
}

@inproceedings{giancola2018soccernet,
  title={Soccernet: A scalable dataset for action spotting in soccer videos},
  author={Giancola, Silvio and Amine, Mohieddine and Dghaily, Tarek and Ghanem, Bernard},
  booktitle={Proceedings of the IEEE conference on computer vision and pattern recognition workshops},
  pages={1711--1721},
  year={2018}
}

@article{wang2024tacticai,
  title={TacticAI: an AI assistant for football tactics},
  author={Wang, Zhe and Veli{\v{c}}kovi{\'c}, Petar and Hennes, Daniel and Toma{\v{s}}ev, Nenad and Prince, Laurel and Kaisers, Michael and Bachrach, Yoram and Elie, Romuald and Wenliang, Li Kevin and Piccinini, Federico and others},
  journal={Nature communications},
  volume={15},
  number={1},
  pages={1906},
  year={2024},
  publisher={Nature Publishing Group UK London}
}

@article{yang2025sga,
  title={SGA-INTERACT: A 3D Skeleton-based Benchmark for Group Activity Understanding in Modern Basketball Tactic},
  author={Yang, Yuchen and Wang, Wei and Liu, Yifei and Dong, Linfeng and Wu, Hao and Zhang, Mingxin and Zhong, Zhihang and Sun, Xiao},
  journal={arXiv preprint arXiv:2503.06522},
  year={2025}
}

@article{dong2024lucidaction,
  title={Lucidaction: A hierarchical and multi-model dataset for comprehensive action quality assessment},
  author={Dong, Linfeng and Wang, Wei and Qiao, Yu and Sun, Xiao},
  journal={Advances in neural information processing systems},
  volume={37},
  pages={96468--96482},
  year={2024}
}

@inproceedings{ibrahim2016hierarchical,
  title={A hierarchical deep temporal model for group activity recognition},
  author={Ibrahim, Mostafa S and Muralidharan, Srikanth and Deng, Zhiwei and Vahdat, Arash and Mori, Greg},
  booktitle={Proceedings of the IEEE conference on computer vision and pattern recognition},
  pages={1971--1980},
  year={2016}
}

@inproceedings{xia2024sportqa,
  title={Sportqa: A benchmark for sports understanding in large language models},
  author={Xia, Haotian and Yang, Zhengbang and Wang, Yuqing and Tracy, Rhys and Zhao, Yun and Huang, Dongdong and Chen, Zezhi and Zhu, Yan and Wang, Yuan-fang and Shen, Weining},
  booktitle={Proceedings of the 2024 Conference of the North American Chapter of the Association for Computational Linguistics: Human Language Technologies (Volume 1: Long Papers)},
  pages={5061--5081},
  year={2024}
}

@inproceedings{mkhallati2023soccernet,
  title={SoccerNet-caption: Dense video captioning for soccer broadcasts commentaries},
  author={Mkhallati, Hassan and Cioppa, Anthony and Giancola, Silvio and Ghanem, Bernard and Van Droogenbroeck, Marc},
  booktitle={Proceedings of the IEEE/CVF Conference on Computer Vision and Pattern Recognition},
  pages={5074--5085},
  year={2023}
}

@article{zou2025deepsport,
  title={DeepSport: A Multimodal Large Language Model for Comprehensive Sports Video Reasoning via Agentic Reinforcement Learning},
  author={Zou, Junbo and Xia, Haotian and Ye, Zhen and Zhang, Shengjie and Lai, Christopher and Ordonez, Vicente and Shen, Weining and Chen, Hanjie},
  journal={arXiv preprint arXiv:2511.12908},
  year={2025}
}

@inproceedings{rao2025multi,
  title={Multi-agent system for comprehensive soccer understanding},
  author={Rao, Jiayuan and Li, Zifeng and Wu, Haoning and Zhang, Ya and Wang, Yanfeng and Xie, Weidi},
  booktitle={Proceedings of the 33rd ACM International Conference on Multimedia},
  pages={3654--3663},
  year={2025}
}

@article{xi2025simple,
  title={A simple yet effective knowledge guided method for entity-aware video captioning on a basketball benchmark},
  author={Xi, Zeyu and Shi, Ge and Li, Xuefen and Yan, Junchi and Li, Zun and Wu, Lifang and Liu, Zilin and Wang, Liang},
  journal={Neurocomputing},
  volume={619},
  pages={129177},
  year={2025},
  publisher={Elsevier}
}

@inproceedings{xi2025player,
  title={Player-centric multimodal prompt generation for large language model based identity-aware basketball video captioning},
  author={Xi, Zeyu and Sun, Haoying and Wu, Yaofei and Yan, Junchi and Zhang, Haoran and Wu, Lifang and Wang, Liang and Chen, Changwen},
  booktitle={Proceedings of the IEEE/CVF International Conference on Computer Vision},
  pages={24330--24339},
  year={2025}
}

@inproceedings{chen2024spatialvlm,
  title={Spatialvlm: Endowing vision-language models with spatial reasoning capabilities},
  author={Chen, Boyuan and Xu, Zhuo and Kirmani, Sean and Ichter, Brain and Sadigh, Dorsa and Guibas, Leonidas and Xia, Fei},
  booktitle={Proceedings of the IEEE/CVF Conference on Computer Vision and Pattern Recognition},
  pages={14455--14465},
  year={2024}
}

@inproceedings{du2024embspatial,
  title={Embspatial-bench: Benchmarking spatial understanding for embodied tasks with large vision-language models},
  author={Du, Mengfei and Wu, Binhao and Li, Zejun and Huang, Xuan-Jing and Wei, Zhongyu},
  booktitle={Proceedings of the 62nd Annual Meeting of the Association for Computational Linguistics (Volume 2: Short Papers)},
  pages={346--355},
  year={2024}
}

@article{team2025gemini,
  title={Gemini robotics: Bringing ai into the physical world},
  author={Team, Gemini Robotics and Abeyruwan, Saminda and Ainslie, Joshua and Alayrac, Jean-Baptiste and Arenas, Montserrat Gonzalez and Armstrong, Travis and Balakrishna, Ashwin and Baruch, Robert and Bauza, Maria and Blokzijl, Michiel and others},
  journal={arXiv preprint arXiv:2503.20020},
  year={2025}
}

@inproceedings{song2025robospatial,
  title={Robospatial: Teaching spatial understanding to 2d and 3d vision-language models for robotics},
  author={Song, Chan Hee and Blukis, Valts and Tremblay, Jonathan and Tyree, Stephen and Su, Yu and Birchfield, Stan},
  booktitle={Proceedings of the Computer Vision and Pattern Recognition Conference},
  pages={15768--15780},
  year={2025}
}

@inproceedings{daxberger2025mm,
  title={Mm-spatial: Exploring 3d spatial understanding in multimodal llms},
  author={Daxberger, Erik and Wenzel, Nina and Griffiths, David and Gang, Haiming and Lazarow, Justin and Kohavi, Gefen and Kang, Kai and Eichner, Marcin and Yang, Yinfei and Dehghan, Afshin and others},
  booktitle={Proceedings of the IEEE/CVF International Conference on Computer Vision},
  pages={7395--7408},
  year={2025}
}

@article{yang2025visual,
  title={Visual spatial tuning},
  author={Yang, Rui and Zhu, Ziyu and Li, Yanwei and Huang, Jingjia and Yan, Shen and Zhou, Siyuan and Liu, Zhe and Li, Xiangtai and Li, Shuangye and Wang, Wenqian and others},
  journal={arXiv preprint arXiv:2511.05491},
  year={2025}
}

@article{wu2025spatialscore,
  title={Spatialscore: Towards unified evaluation for multimodal spatial understanding},
  author={Wu, Haoning and Huang, Xiao and Chen, Yaohui and Zhang, Ya and Wang, Yanfeng and Xie, Weidi},
  journal={arXiv e-prints},
  pages={arXiv--2505},
  year={2025}
}

@article{yang2025cambrian,
  title={Cambrian-s: Towards spatial supersensing in video},
  author={Yang, Shusheng and Yang, Jihan and Huang, Pinzhi and Brown, Ellis and Yang, Zihao and Yu, Yue and Tong, Shengbang and Zheng, Zihan and Xu, Yifan and Wang, Muhan and others},
  journal={arXiv preprint arXiv:2511.04670},
  year={2025}
}

@inproceedings{
zhang2026on,
title={On the Generalization Capacities of {MLLM}s for Spatial Intelligence},
author={Gongjie Zhang and Wenhao Li and Quanhao Qian and Jiuniu Wang and Deli Zhao and Shijian Lu and Ran Xu},
booktitle={The Fourteenth International Conference on Learning Representations},
year={2026}
}

@article{wu2025spatial,
  title={Spatial-mllm: Boosting mllm capabilities in visual-based spatial intelligence},
  author={Wu, Diankun and Liu, Fangfu and Hung, Yi-Hsin and Duan, Yueqi},
  journal={arXiv preprint arXiv:2505.23747},
  year={2025}
}

@article{ouyang2025spacer,
  title={Spacer: Reinforcing mllms in video spatial reasoning},
  author={Ouyang, Kun and Liu, Yuanxin and Wu, Haoning and Liu, Yi and Zhou, Hao and Zhou, Jie and Meng, Fandong and Sun, Xu},
  journal={arXiv preprint arXiv:2504.01805},
  year={2025}
}

@inproceedings{dai2017scannet,
  title={Scannet: Richly-annotated 3d reconstructions of indoor scenes},
  author={Dai, Angela and Chang, Angel X and Savva, Manolis and Halber, Maciej and Funkhouser, Thomas and Nie{\ss}ner, Matthias},
  booktitle={Proceedings of the IEEE conference on computer vision and pattern recognition},
  pages={5828--5839},
  year={2017}
}

@inproceedings{ling2024dl3dv,
  title={Dl3dv-10k: A large-scale scene dataset for deep learning-based 3d vision},
  author={Ling, Lu and Sheng, Yichen and Tu, Zhi and Zhao, Wentian and Xin, Cheng and Wan, Kun and Yu, Lantao and Guo, Qianyu and Yu, Zixun and Lu, Yawen and others},
  booktitle={Proceedings of the IEEE/CVF Conference on Computer Vision and Pattern Recognition},
  pages={22160--22169},
  year={2024}
}

@article{deng2025internspatial,
  title={Internspatial: A comprehensive dataset for spatial reasoning in vision-language models},
  author={Deng, Nianchen and Gu, Lixin and Ye, Shenglong and He, Yinan and Chen, Zhe and Li, Songze and Wang, Haomin and Wei, Xingguang and Yang, Tianshuo and Dou, Min and others},
  journal={arXiv preprint arXiv:2506.18385},
  year={2025}
}

@article{mmsi-video,
  title={MMSI-Video-Bench: A Holistic Benchmark for Video-Based Spatial Intelligence},
  author={Lin, Jingli and Xu, Runsen and Zhu, Shaohao and Yang, Sihan and Cao, Peizhou and Ran, Yunlong and Hu, Miao and Zhu, Chenming and Xie, Yiman and Long, Yilin and others},
  journal={arXiv preprint arXiv:2512.10863},
  year={2025}
}

@article{zheng2025learning,
  title={Learning from videos for 3d world: Enhancing mllms with 3d vision geometry priors},
  author={Zheng, Duo and Huang, Shijia and Li, Yanyang and Wang, Liwei},
  journal={arXiv preprint arXiv:2505.24625},
  year={2025}
}

@inproceedings{SITE,
  title={Site: towards spatial intelligence thorough evaluation},
  author={Wang, Wenqi and Tan, Reuben and Zhu, Pengyue and Yang, Jianwei and Yang, Zhengyuan and Wang, Lijuan and Kolobov, Andrey and Gao, Jianfeng and Gong, Boqing},
  booktitle={Proceedings of the IEEE/CVF International Conference on Computer Vision},
  pages={9058--9069},
  year={2025}
}

@article{li2025viewspatial,
  title={Viewspatial-bench: Evaluating multi-perspective spatial localization in vision-language models},
  author={Li, Dingming and Li, Hongxing and Wang, Zixuan and Yan, Yuchen and Zhang, Hang and Chen, Siqi and Hou, Guiyang and Jiang, Shengpei and Zhang, Wenqi and Shen, Yongliang and others},
  journal={arXiv preprint arXiv:2505.21500},
  year={2025}
}

@article{sun2025spacevista,
  title={Spacevista: All-scale visual spatial reasoning from mm to km},
  author={Sun, Peiwen and Lang, Shiqiang and Wu, Dongming and Ding, Yi and Feng, Kaituo and Liu, Huadai and Ye, Zhen and Liu, Rui and Liu, Yun-Hui and Wang, Jianan and others},
  journal={arXiv preprint arXiv:2510.09606},
  year={2025}
}

@inproceedings{
cai2026depthlm,
title={Depth{LM}: Metric Depth from Vision Language Models},
author={Zhipeng Cai and Ching-Feng Yeh and Hu Xu and Zhuang Liu and Gregory P. Meyer and Xinjie Lei and Changsheng Zhao and Shang-Wen Li and Vikas Chandra and Yangyang Shi},
booktitle={The Fourteenth International Conference on Learning Representations},
year={2026}
}

@inproceedings{
li2026spatialladder,
title={SpatialLadder: Progressive Training for Spatial Reasoning in Vision-Language Models},
author={Hongxing Li and Dingming Li and Zixuan Wang and Yuchen Yan and Hang Wu and Wenqi Zhang and Yongliang Shen and Weiming Lu and Jun Xiao and Yueting Zhuang},
booktitle={The Fourteenth International Conference on Learning Representations},
year={2026}
}

@article{zhu2023tame,
  title={Tame a wild camera: In-the-wild monocular camera calibration},
  author={Zhu, Shengjie and Kumar, Abhinav and Hu, Masa and Liu, Xiaoming},
  journal={Advances in Neural Information Processing Systems},
  volume={36},
  pages={45137--45149},
  year={2023}
}

@article{lin2025depth,
  title={Depth anything 3: Recovering the visual space from any views},
  author={Lin, Haotong and Chen, Sili and Liew, Junhao and Chen, Donny Y and Li, Zhenyu and Shi, Guang and Feng, Jiashi and Kang, Bingyi},
  journal={arXiv preprint arXiv:2511.10647},
  year={2025}
}

@inproceedings{van20223d,
  title={3D ball localization from a single calibrated image},
  author={Van Zandycke, Gabriel and De Vleeschouwer, Christophe},
  booktitle={Proceedings of the IEEE/CVF Conference on Computer Vision and Pattern Recognition},
  pages={3472--3480},
  year={2022}
}

@article{carion2025sam,
  title={Sam 3: Segment anything with concepts},
  author={Carion, Nicolas and Gustafson, Laura and Hu, Yuan-Ting and Debnath, Shoubhik and Hu, Ronghang and Suris, Didac and Ryali, Chaitanya and Alwala, Kalyan Vasudev and Khedr, Haitham and Huang, Andrew and others},
  journal={arXiv preprint arXiv:2511.16719},
  year={2025}
}

@article{wang2025internvl35,
  title={Internvl3. 5: Advancing open-source multimodal models in versatility, reasoning, and efficiency},
  author={Wang, Weiyun and Gao, Zhangwei and Gu, Lixin and Pu, Hengjun and Cui, Long and Wei, Xingguang and Liu, Zhaoyang and Jing, Linglin and Ye, Shenglong and Shao, Jie and others},
  journal={arXiv preprint arXiv:2508.18265},
  year={2025}
}

@article{team2025kimi,
  title={Kimi-vl technical report},
  author={Team, Kimi and Du, Angang and Yin, Bohong and Xing, Bowei and Qu, Bowen and Wang, Bowen and Chen, Cheng and Zhang, Chenlin and Du, Chenzhuang and Wei, Chu and others},
  journal={arXiv preprint arXiv:2504.07491},
  year={2025}
}

@article{an2025llava,
  title={Llava-onevision-1.5: Fully open framework for democratized multimodal training},
  author={An, Xiang and Xie, Yin and Yang, Kaicheng and Zhang, Wenkang and Zhao, Xiuwei and Cheng, Zheng and Wang, Yirui and Xu, Songcen and Chen, Changrui and Zhu, Didi and others},
  journal={arXiv preprint arXiv:2509.23661},
  year={2025}
}

@article{li2024llava,
  title={Llava-onevision: Easy visual task transfer},
  author={Li, Bo and Zhang, Yuanhan and Guo, Dong and Zhang, Renrui and Li, Feng and Zhang, Hao and Zhang, Kaichen and Zhang, Peiyuan and Li, Yanwei and Liu, Ziwei and others},
  journal={arXiv preprint arXiv:2408.03326},
  year={2024}
}

@article{zhu2025internvl3,
  title={Internvl3: Exploring advanced training and test-time recipes for open-source multimodal models},
  author={Zhu, Jinguo and Wang, Weiyun and Chen, Zhe and Liu, Zhaoyang and Ye, Shenglong and Gu, Lixin and Tian, Hao and Duan, Yuchen and Su, Weijie and Shao, Jie and others},
  journal={arXiv preprint arXiv:2504.10479},
  year={2025}
}

@inproceedings{zheng2024llamafactory,
  title={Llamafactory: Unified efficient fine-tuning of 100+ language models},
  author={Zheng, Yaowei and Zhang, Richong and Zhang, Junhao and Ye, Yanhan and Luo, Zheyan},
  booktitle={Proceedings of the 62nd annual meeting of the association for computational linguistics (volume 3: system demonstrations)},
  pages={400--410},
  year={2024}
}

@inproceedings{kienzle2026uplifting,
  title={Uplifting Table Tennis: A Robust, Real-World Application for 3D Trajectory and Spin Estimation},
  author={Kienzle, Daniel and Ludwig, Katja and Lorenz, Julian and Satoh, {Shin'ichi} and Lienhart, Rainer},
  booktitle={Proceedings of the IEEE/CVF Winter Conference on Applications of Computer Vision (WACV)},
  year={2026}
}

@inproceedings{wildcamera,
  title     = {Tame a Wild Camera: In-the-Wild Monocular Camera Calibration},
  author    = {Shen, Shengjie and Hua, Yunzhi and Jiang, Zhipeng and Li, Zhicheng and Gao, Mingze and Ge, Zequn and Han, Zhiqiang and Zhong, Fan and Chen, Xinggang},
  booktitle = {Advances in Neural Information Processing Systems},
  year      = {2023}
}

@article{gao2026holispatial,
      title={Holi-Spatial: Evolving Video Streams into Holistic 3D Spatial Intelligence}, 
      author={Yuanyuan Gao and Hao Li and Yifei Liu and Xinhao Ji and Yuning Gong and Yuanjun Liao and Fangfu Liu and Manyuan Zhang and Yuchen Yang and Dan Xu and Xue Yang and Huaxi Huang and Hongjie Zhang and Ziwei Liu and Xiao Sun and Dingwen Zhang and Zhihang Zhong},
      journal={arXiv preprint arXiv:2603.07660},
      year={2026}
}
}

\clearpage
\setcounter{page}{1}
\appendix

\section*{Contents}
\noindent
A. \hspace{0.2cm} Data Engine Details \dotfill \pageref{sec:supp-pipline-details} \\
B. \hspace{0.2cm} CourtSI Details \dotfill \pageref{sec:supp-courtsi} \\
C. \hspace{0.2cm} Experiment Details \dotfill \pageref{sec:supp-exp} \\
D. \hspace{0.2cm} Ethical Considerations \dotfill \pageref{sec:supp-ethical-cons} \\

\section{Data Engine Details}
\subsection{Pipeline Details}
\label{sec:supp-pipline-details}
The proposed data engine comprises court annotation, ball annotation, and player mesh recovery. We detail each in the following.

\paragraph{\textbf{Court Annotation}.} We estimate the camera parameters using a PnP solver with annotated court keypoints and their corresponding 3D positions derived from the fixed court geometry. We develop an interactive panel to assist annotators in selecting keypoints. 

The annotation is performed on the raw videos, and a frame in which all court keypoints are clearly visible is selected for calibration.
For scenes with a static camera view, the estimated calibration parameters are reused across all frames. 
For a few cases with dynamic camera views, we propagate the court keypoints to adjacent frames and apply the proposed calibration method to estimate the camera parameters of each neighboring frame using the transformed keypoints.
Specifically, we utilize DepthAnythingV3 to estimate both per-pixel depth and relative camera parameters between frames. Given the annotated reference frame and an adjacent frame, it provides the depth map as well as the camera intrinsics and extrinsics for each frame.
Using the estimated depth, 2D keypoints in the reference frame are first back-projected into 3D space. The 3D points are then transformed to the coordinate system of the adjacent frame using the relative camera pose, and finally re-projected onto the image plane of the adjacent frame to obtain the corresponding 2D positions.

With reprojection-based verification, we observe that this propagation process is effective across frames.
Although DepthAnythingV3 supports metric depth estimation and multi-frame camera parameter estimation, we find that directly using the calibrated reference frame as input and relying on DepthAnythingV3 to propagate camera parameters to subsequent frames introduces significant calibration errors. In practice, the accumulated pose and scale inconsistencies lead to noticeable reprojection deviations. Therefore, instead of directly adopting the propagated camera parameters, we use DepthAnythingV3 primarily for depth-guided geometric transfer and perform camera parameter estimation independently to ensure calibration stability.

\begin{figure}[t]
    \centering
    \includegraphics[width=\linewidth]{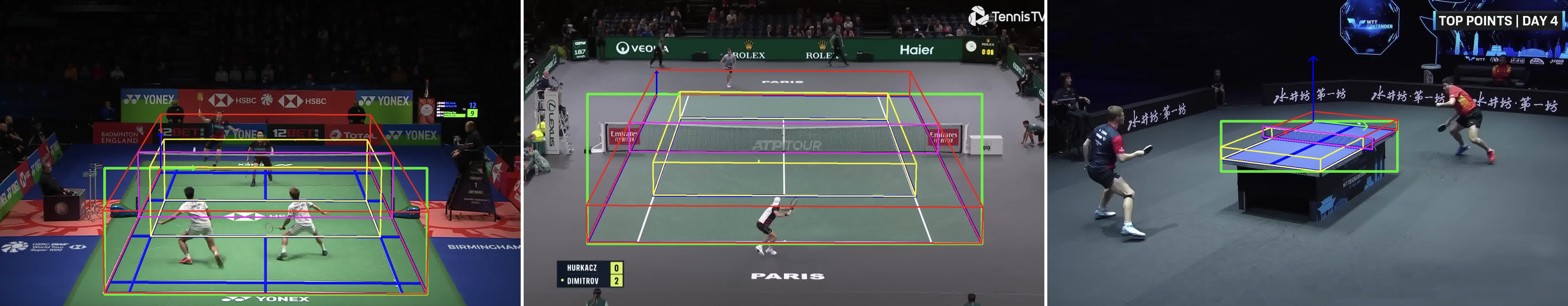}
    \caption{Calibration examples. A 3D court box with real-world dimensions is reprojected onto the image using the estimated camera parameters. The close alignment between the projected court structure and the image clues indicates strong reprojection consistency, validating the accuracy of the calibration.}
    \label{fig:court-annot}
\end{figure}

The calibration results are illustrated in \cref{fig:court-annot}.
The world coordinate system is defined in a right-handed format as follows: the origin is located at the far corner point of the court from the camera’s perspective; the x-axis is aligned with the court length and is positive toward the camera; the y-axis is aligned with the court width and is positive toward the camera; and the z-axis is perpendicular to the court plane.

\begin{figure}[t]
    \centering
    \includegraphics[width=\linewidth]{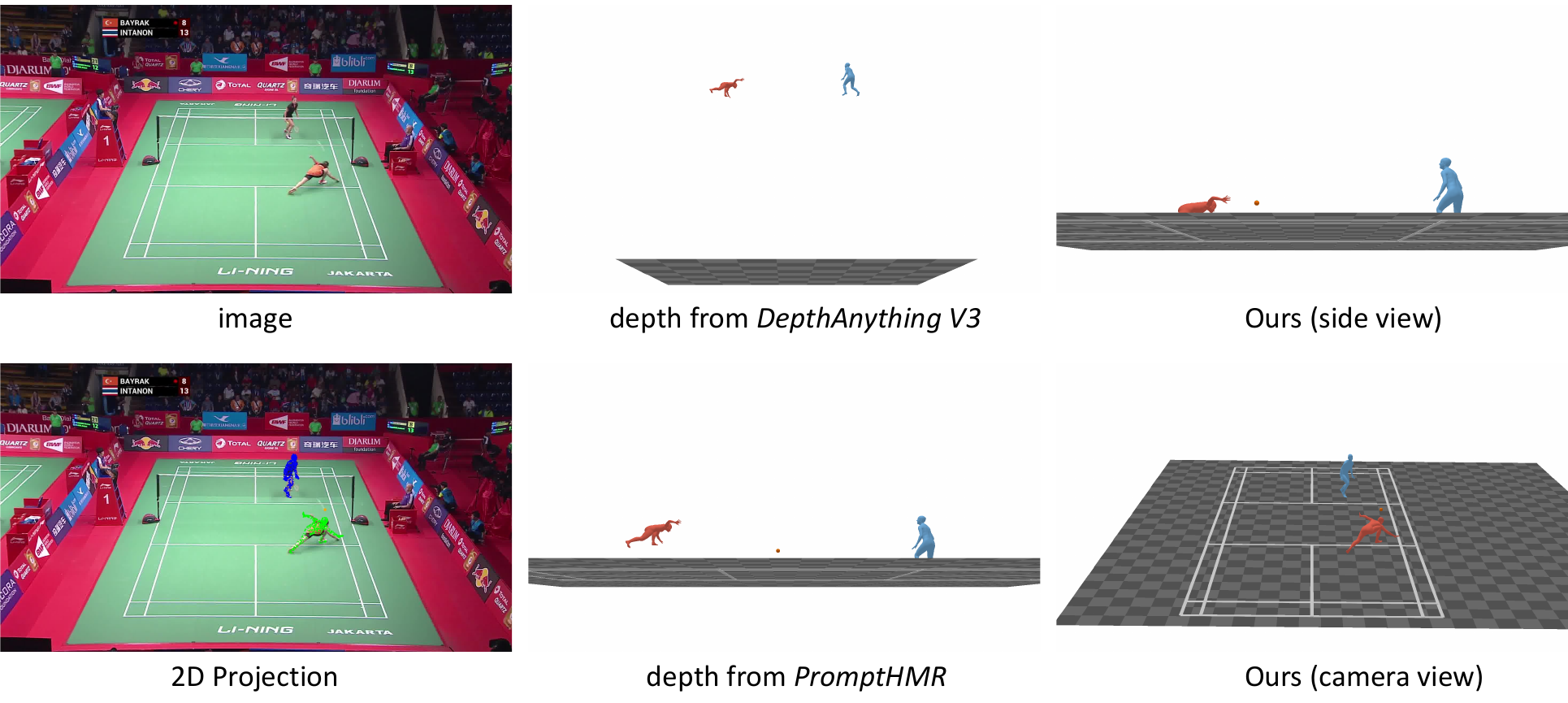}
    \caption{Illustration of player depth estimation using PromptHMR, DepthAnythingV3, and our method. All baselines take metric-scale camera intrinsics as input.}
    \label{fig:supp-player}
\end{figure}

\paragraph{\textbf{Ball Annotation}.}
In the main text, we describe the ball annotation process within a single frame by converting depth estimation into ball projection estimation on the court ground plane.
For the raw video data, we model the ball trajectory during each rally. While the ball is airborne, until it is struck by a player or contacts the court, it is primarily influenced by gravity, aerodynamic lift generated by spin, and air resistance. We approximate this motion as a constant-acceleration problem. Annotators label the start point, midpoint, and end point of each trajectory segment, from which the 3D acceleration and initial velocity can be estimated.
This approach significantly reduces the annotation effort required for airborne ball tracking. When the estimated trajectory does not meet quality standards, we revert to per-frame annotation to ensure accuracy.
For table tennis, we adopt the 2D-to-3D lifting approach proposed in~\cite{kienzle2026uplifting} to estimate the ball position. This method takes the court corner positions as input and is trained on large-scale, high-fidelity simulation data, which enhances robustness under our experimental conditions.

\paragraph{\textbf{Player Mesh Recovery}.}
PromptHMR estimates the human mesh in camera coordinates, conditioned on the bounding box and camera parameters.
We first employ SAM3 to track target players using the prompt “player.” Although SAM3 performs well in most cases, we develop an interactive refinement panel to manually correct a small number of inaccurate detections.
Regarding camera parameters, PromptHMR assumes simplified camera intrinsics, where the focal lengths along the x- and y-axes are identical and the principal point is located at the image center. We therefore optimize this simplified camera model using the previously annotated court keypoints. Based on quality control, this simplification does not involve much error in final localization.

As discussed in the main text, we observe that the estimated depths of the recovered human meshes are often inaccurate. \cref{fig:supp-player} provides an qualitative example.
To address this issue, annotators manually estimate the depth of the lowest mesh vertex using the same strategy as in ball annotation, by labeling its height above the court surface. The entire mesh is then re-aligned according to the corrected depth.
Instead of directly translating the mesh by the depth offset, which would distort its 3D scale, we apply a similarity transformation centered at the camera location $C$:
\begin{equation}
    X' = sX + (1 - s) C,
\end{equation}
where the depth scale factor $s$ is computed from the depth correction of the lowest vertex. This transformation uniformly rescales the mesh along rays emanating from the camera center: the mesh is enlarged when the corrected depth is closer to the camera (i.e., smaller depth), and shrunk when the corrected depth is farther away (i.e., larger depth).

\subsection{Comparison with Monocular Scene Reconstruction Methods}
\label{sec:supp-compare-mono-methods}

In this section, we present a detailed comparison with monocular scene reconstruction methods using the multi-view evaluation dataset introduced in \cref{sec:data-engine}.

\begin{table}[t]
\centering
\caption{Quantitative error analysis of camera intrinsic parameters.}
\begin{tabular}{l cc cc cc}
\toprule
\multirow{2}{*}{\textbf{Baseline}} 
  & \multicolumn{2}{c}{\textbf{Badminton}} 
  & \multicolumn{2}{c}{\textbf{Tennis}} 
  & \multicolumn{2}{c}{\textbf{Table Tennis}} \\
\cmidrule(lr){2-3} \cmidrule(lr){4-5} \cmidrule(lr){6-7}
& $e_{fx}$(\%) & $e_{fy}$(\%) 
& $e_{fx}$(\%) & $e_{fy}$(\%) 
& $e_{fx}$(\%) & $e_{fy}$(\%) \\
\midrule
WildCamera      & 67.20 & 70.86 & 17.48 & 12.16 & 13.66 & 15.20 \\
DepthAnythingV3 & 38.89 & 38.24 &  5.42 &  4.16 &  0.85 &  0.90 \\
Ours            &  0.55 &  0.72 &  4.23 &  4.31 &  0.01 &  1.45 \\
\bottomrule
\end{tabular}
\label{tab:focal-length-error}
\end{table}

\paragraph{\textbf{Camera Calibration}.}
Following~\cite{wildcamera}, we measure the accuracy of the estimated camera intrinsics
using the relative focal length errors $e_{f_x}$ and $e_{f_y}$, defined as:
\begin{equation}
    e_{f_x} = \frac{|f_x^{\text{pred}} - f_x^{\text{gt}}|}{f_x^{\text{gt}}}, \quad
    e_{f_y} = \frac{|f_y^{\text{pred}} - f_y^{\text{gt}}|}{f_y^{\text{gt}}}.
\end{equation}
Both metrics are reported as percentages and averaged over all video sequences
within each sport category, as shown in \cref{tab:focal-length-error}.
The results indicate that our calibration method achieves the best performance across the three sports scenarios by explicitly leveraging court geometry.

\begin{figure}[t]
    \centering
    \includegraphics[width=\linewidth]{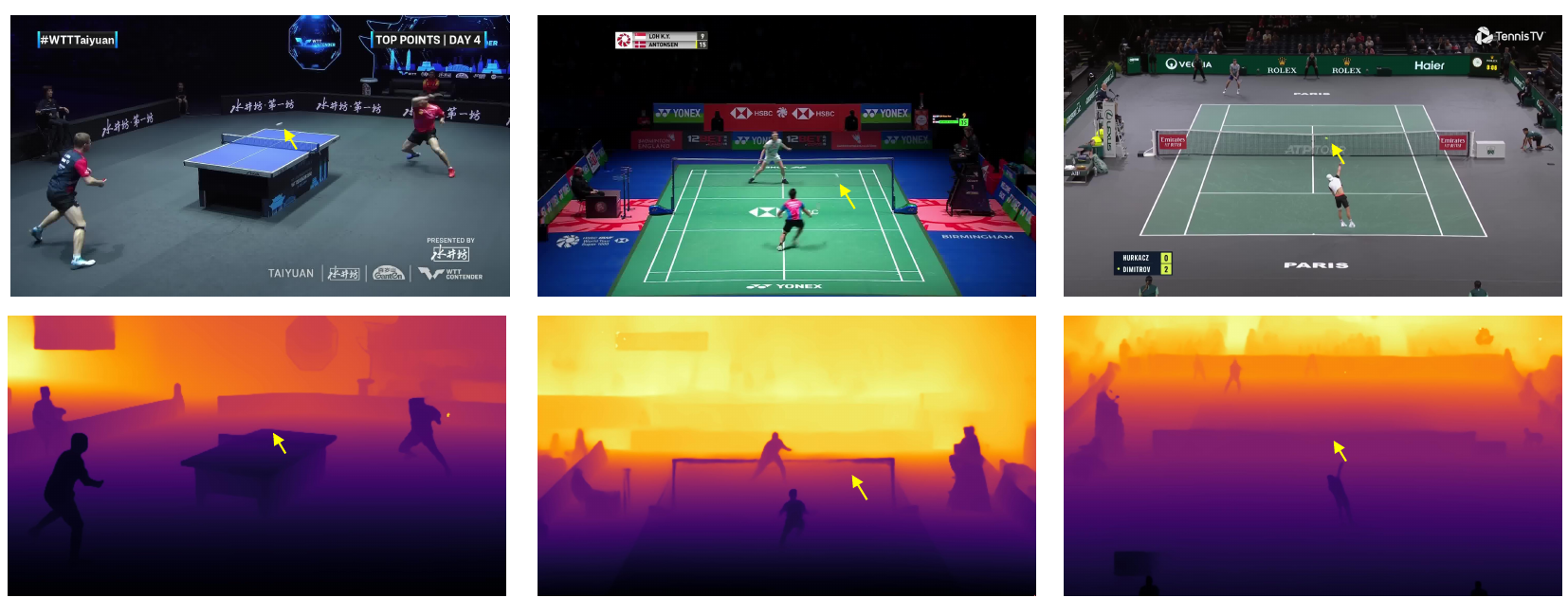}
    \caption{Original broadcast frames (top) and depth maps estimated by DepthAnythingV3 (bottom). Yellow arrows indicate the ball positions.}
    \label{fig:supp-depth}
\end{figure}

\begin{table}[t]
\centering
\caption{Quantitative error analysis of ball localization. $^*$ denotes using ground truth 2D ball locations and camera parameters.}
\resizebox{\linewidth}{!}{
\begin{tabular}{l ccc ccc ccc}
\toprule
\multirow{2}{*}{\textbf{Baseline}} 
  & \multicolumn{3}{c}{\textbf{Badminton}} 
  & \multicolumn{3}{c}{\textbf{Tennis}} 
  & \multicolumn{3}{c}{\textbf{Table Tennis}} \\
\cmidrule(lr){2-4} \cmidrule(lr){5-7} \cmidrule(lr){8-10}
& {$X$} & {$Y$} & {$Z$} 
& {$X$} & {$Y$} & {$Z$} 
& {$X$} & {$Y$} & {$Z$} \\
\midrule
{DepthAnythingV3$^*$} 
  & $1227\mathrm{cm}$ & $252\mathrm{cm}$ & $241\mathrm{cm}$
  & $3168\mathrm{cm}$ & $2045\mathrm{cm}$ & $1833\mathrm{cm}$
  & $249\mathrm{cm}$ & $199\mathrm{cm}$ & $26\mathrm{cm}$ \\
{Ours} 
  & $10\mathrm{cm}$ & $4\mathrm{cm}$ & $6\mathrm{cm}$
  & $29\mathrm{cm}$ & $11\mathrm{cm}$ & $9\mathrm{cm}$
  & $0.3\mathrm{cm}$ & $0.3\mathrm{cm}$ & $0.5\mathrm{cm}$ \\
\bottomrule
\end{tabular}}
\label{tab:ball-localization}
\end{table}

\paragraph{\textbf{Ball Localization}.}
For ball localization, we employ DepthAnythingV3 as the metric depth estimator, following a two-stage pipeline of detection followed by lifting.
As shown in \cref{tab:ball-localization}, even when provided with ground-truth 2D locations and camera parameters, DepthAnythingV3 fails to produce accurate metric depth estimates. Qualitatively, as illustrated in \cref{fig:supp-depth}, this failure likely arises because the ball occupies only a small image region, and the predicted depth map cannot capture such subtle details.

\begin{table}[t]
\centering
\caption{Quantitative error analysis of player localization.}
\begin{tabular}{l @{\hspace{5pt}} c @{\hspace{5pt}} c @{\hspace{5pt}} c @{\hspace{5pt}} c}
\toprule
\multirow{2}{*}{\textbf{Baseline}} & \multicolumn{3}{c}{\textbf{Pelvis}} \\
\cmidrule(lr){2-4}
{} & {Badminton} & {Tennis} & {Table Tennis} \\
\midrule
{PromptHMR} & {$134\mathrm{cm}$} & {$1144\mathrm{cm}$} & {$33\mathrm{cm}$} \\
{DepthAnythingV3} & {$2191\mathrm{cm}$} & {$2744\mathrm{cm}$} & {$62\mathrm{cm}$}\\
{Ours} & {$21\mathrm{cm}$} & {$27\mathrm{cm}$} & {$16\mathrm{cm}$}\\
\bottomrule
\end{tabular}
\label{tab:supp-player-quality}
\end{table}

\paragraph{\textbf{Player Localization}.}
In \cref{tab:supp-player-quality}, we report the quantitative evaluation of player localization. For all baseline methods, the estimated camera parameters are used as input, and the predicted depth is employed to transform the human mesh into the 3D world coordinate system. We then compute the 3D pelvis position error as the localization metric. The results demonstrate that our localization method outperforms the baseline approaches.

\section{CourtSI Details}
\label{sec:supp-courtsi}

\subsection{Question Template}
\label{sec:supp-question-template}
All query templates evaluated in our benchmark follow a systematic structure composed of three components: pre-prompt, question, and post-prompt.
Specifically, each query is formed by combining a pre-prompt, a question, and a post-prompt. The diversity of queries in our benchmark primarily arises from the use of different question types.

\paragraph{\textbf{Pre-prompt}} establishes the contextual background of the problem and mitigates any potential ambiguities inherent in the query. Its standardized content is defined as follows:
\begin{prompt}
This is a snapshot from a \{\textit{sport name}\} match view from a high angle. The court closer to the camera is the `near court', and the opposite one is the `far court'. All references to `left' or `right' in the questions describing the court or relative positions are based on the camera's perspective, corresponding to the left and right sides of the image frame. However, references to specific body parts (e.g., `left wrist',`right knee') follow the player's anatomical perspective (the player's own left/right).
\end{prompt}

\paragraph{\textbf{Post-prompt}} delineates the explicit formatting rules for the model's output. Depending on the inquiry category, the model must follow one of two high-level output groups: numerical or multiple-choice (MCQ).
Numerical outputs include floating-point numbers, integers, or 3D spatial coordinates, while MCQ output is a single multiple-choice option. Its content is defined as follows:
\begin{prompt}
\textbf{floating-point number}: Answer with a single float number representing meters. Example: 2.54
\smallskip

\textbf{3D coordinate}: Answer strictly in the format (x, y, z) with no units. Example: (1.2, 3.4, 0.0)
\smallskip

\textbf{integer}: Answer with a single integer number. Example: 3
\smallskip

\textbf{multiple-choice option}: Select the best option. Output only the single uppercase letter corresponding to the choice. Example: B

\end{prompt}
\paragraph{\textbf{Question}} can be broadly classified into 13 primary categories. When further categorized by the generated templates, they expand into 20 distinct types, comprising a total of 94 unique templates. In this section, we present a selection of the template generation results, organized according to the classification methodology detailed in the paper.

Within these examples below, bold text denotes the question category. Italicized text indicates interchangeable variables or elements requiring additional specification within the template. Numerical indices represent varied phrasing or content variations within the same question category (an exhaustive list is omitted due to space limitations). Based on our quantitative assessment, this question generation strategy is capable of producing 4,403 entirely distinct questions across three different ball sports.

\begin{prompt}
{\centering \textbf{Distance Measurement} \par}

\textbf{camera-object}:

\smallskip
    1. How far apart are the camera and \textit{object} in 3D space?
    
    2. Calculate the 3D Euclidean distance between the camera and \textit{object} in meters.
\smallskip

\textbf{object-object}:

\smallskip
    1. What is the distance between \textit{object1} and \textit{object2} in meters?
    
    2. If a line were drawn directly from \textit{object1} to \textit{object2}, what would be its length in meters?
\smallskip
    
\textbf{object-line}: 

\smallskip
    1. What is the perpendicular distance from \textit{object} to \textit{line}?
    
    2. Mapping \textit{object}'s position to \textit{the court zone}/\textit{the table surface}, what is its perpendicular distance to \textit{line} in meters?
\smallskip

\textbf{height}:

    1. What is the height of \textit{object} in meters at this moment?
    
    2. How high above the court surface is \textit{object} currently positioned?

\smallskip

{\centering \textbf{Spatial Counting} \par}
\textbf{player}:

\smallskip
1. How many players are visible on the court in this image?

2. Count the total number of players currently playing in the match.
\smallskip

\textbf{ball}:

\smallskip
1. Can you see \textit{the tennis ball}/\textit{the ping pong ball}/\textit{the shuttlecock} in the snapshot?

(A)Yes
(B)No

2. Is \textit{the tennis ball}/\textit{the ping pong ball}/\textit{the shuttlecock} visible in this image?

(A)Yes
(B)No
\smallskip

{\centering \textbf{Localization} \par}
\textbf{Object}: 

\smallskip
    Using a coordinate system where the origin (0,0,0) is \textit{the intersection of the far baseline and the left doubles sideline}/\textit{the top-left corner of the table surface}. The X-axis extends along the sideline towards the camera, the Y-axis extends along \textit{the far baseline}/\textit{the far endline} to the right, and the Z-axis is vertical. 
    
    1. What is the 3D coordinate (x, y, z) of \textit{object} in meters?

    2. Locate \textit{object} within the defined coordinate system and return its (x, y, z) values.

{\centering \textbf{Relational Reasoning} \par}
\smallskip
\textbf{player-player}:

\smallskip
    1. Measuring from the pelvis of each player, which of these players is closest to \textit{player}?
      
    (A)Player 1
    (B)Player 2
    (C)Player 4 \textit{(set options according to the situation)}

    2. Based on \textit{player}'s perspective, is \textit{player} located to their left or right?

    (A)Left side
    (B)Right side

    3. Is \textit{player} positioned to the left or to the right of \textit{player} from the camera's view?
    
    (A)Left
    (B)Right
    (C)Directly in front or behind
    
\smallskip
\textbf{ball-zone}:

\smallskip
1. In which longitudinal zone of the court is the tennis ball currently located?

(A)The forecourt (between the net and the service line)
(B)The midcourt (between the service line and the baseline)
(C)The backcourt (outside the baseline)
        
2. Is the shuttlecock currently positioned above or below the top edge of the net?

(A)Above the net
(B)Below the net

3. Is the ping pong ball on the left or right side of the table center line?

    (A)Left side
    (B)Right side
    (C)On the center line

\smallskip
\textbf{ball-player}:
\smallskip

1. Measuring from the pelvis of each player, which player has the smallest Euclidean distance to \textit{the tennis ball}/\textit{the ping pong ball}/\textit{the shuttlecock}?
      
    (A)Player 1
    (B)Player 2
    (C)Player 4 \textit{(set options according to the situation)}

2. Imagine you are \textit{player}. Is \textit{the tennis ball}/\textit{the ping pong ball}/\textit{the shuttlecock} currently to your left-hand side or right-hand side?

    (A)Left side
    (B)Right side

3. From the camera's perspective, which side is \textit{player} on relative to \textit{the tennis ball}/\textit{the ping pong ball}/\textit{the shuttlecock}?

    (A)Left
    (B)Right
    (C)Directly in front or behind

\smallskip
\textbf{cam-player}:

\smallskip
1. Measuring from the pelvis of each player, which of these players is closest to the camera?

    (A)Player 1
    (B)Player 2
    (C)Player 4 \textit{(set options according to the situation)}

2. From the ego-centric view of \textit{player}, which side is the camera on?

    (A)Left side
    (B)Right side

3. Is \textit{player} positioned to the left or to the right of the camera from the camera's view?
    
    (A)Left
    (B)Right
    (C)Directly in front or behind
 
\smallskip
\textbf{player-zone}:

\smallskip
1. Classify the position of \textit{player} into one of the three court zones: forecourt, midcourt, or backcourt.

(A)The forecourt (between the net and the service line)
(B)The midcourt (between the service line and the baseline)
(C)The backcourt (outside the baseline)

2. Where is {object1} standing relative to the length of the court?

(A)Front court
(B)Mid court
(C)Rear court

\smallskip
\textbf{player-line}:

\smallskip
1. Considering the pelvis position of each player, which player has the smallest perpendicular distance to \textit{line}?

(A)Player 1
(B)Player 2
(C)Player 3
(D)Player 4

2. Based on the pelvis positions, which player is nearest to \textit{line} in terms of perpendicular distance?

(A)Player 1
(B)Player 2
(C)Player 3
(D)Player 4

Is \textit{player} positioned to the left or to the right of \textit{player} from the camera's view?

\end{prompt}

\clearpage
\subsection{QA examples}

\begin{center}
\includegraphics[width=\linewidth,height=0.25\textheight,keepaspectratio]{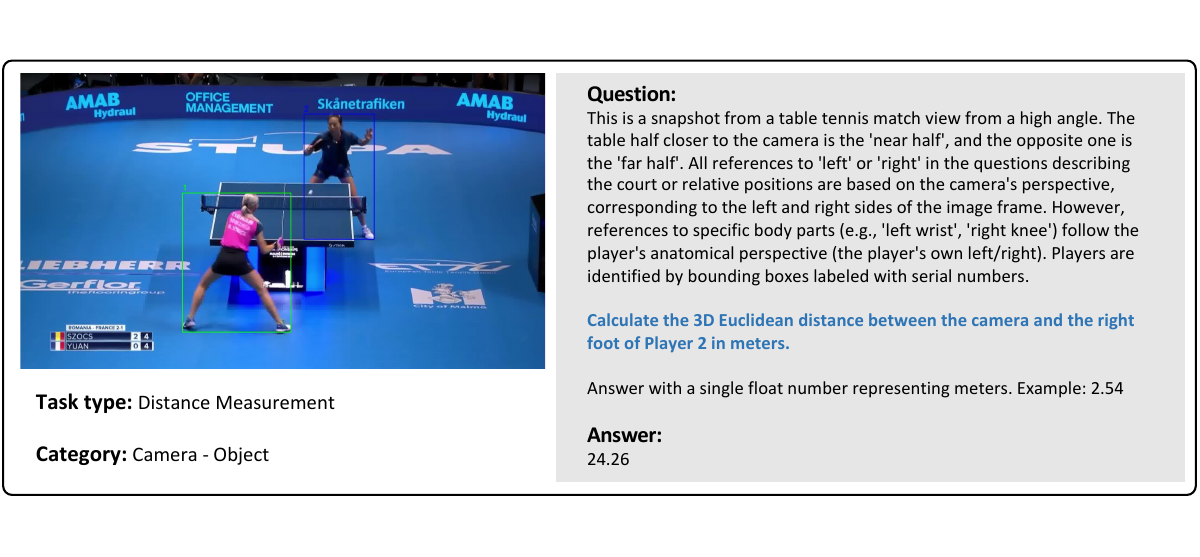}
\end{center}
\vspace{-10mm}

\begin{center}
\includegraphics[width=\linewidth,height=0.25\textheight,keepaspectratio]{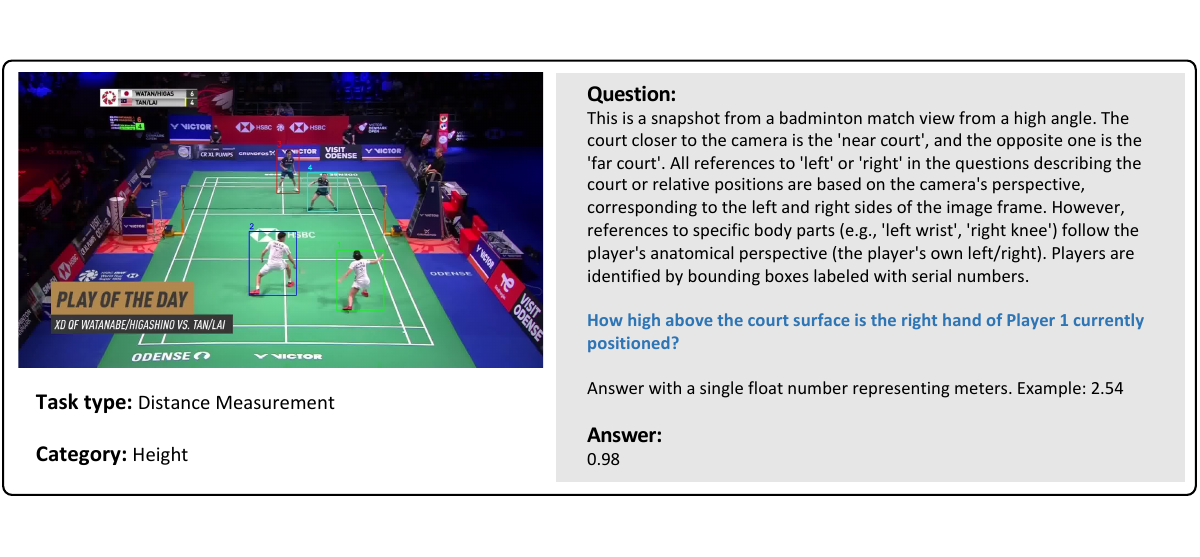}
\end{center}
\vspace{-10mm}

\begin{center}
\includegraphics[width=\linewidth,height=0.25\textheight,keepaspectratio]{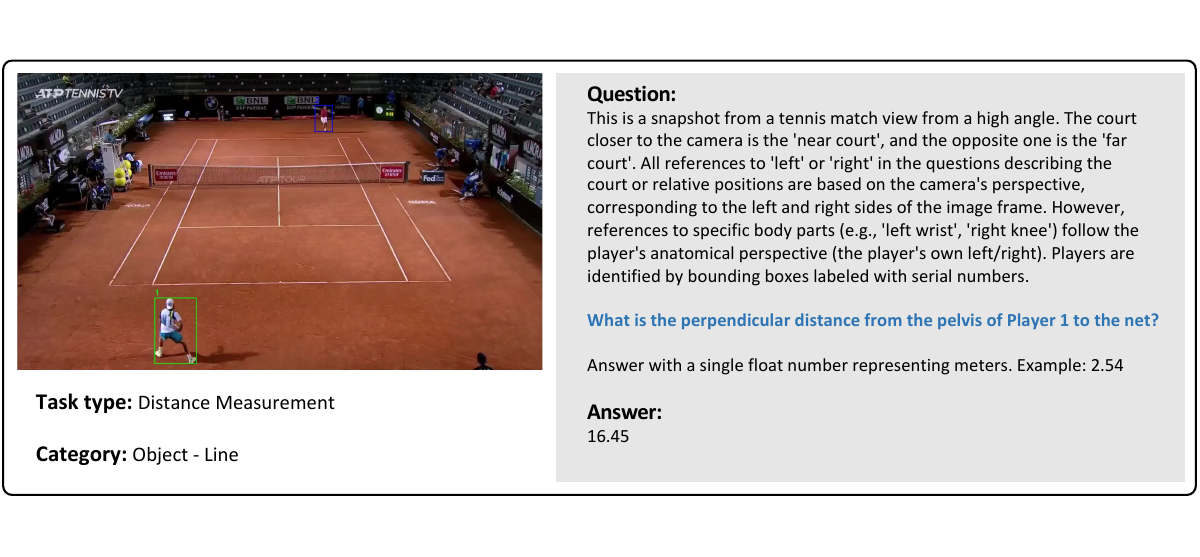}
\end{center}
\vspace{-10mm}

\begin{center}
\includegraphics[width=\linewidth,height=0.25\textheight,keepaspectratio]{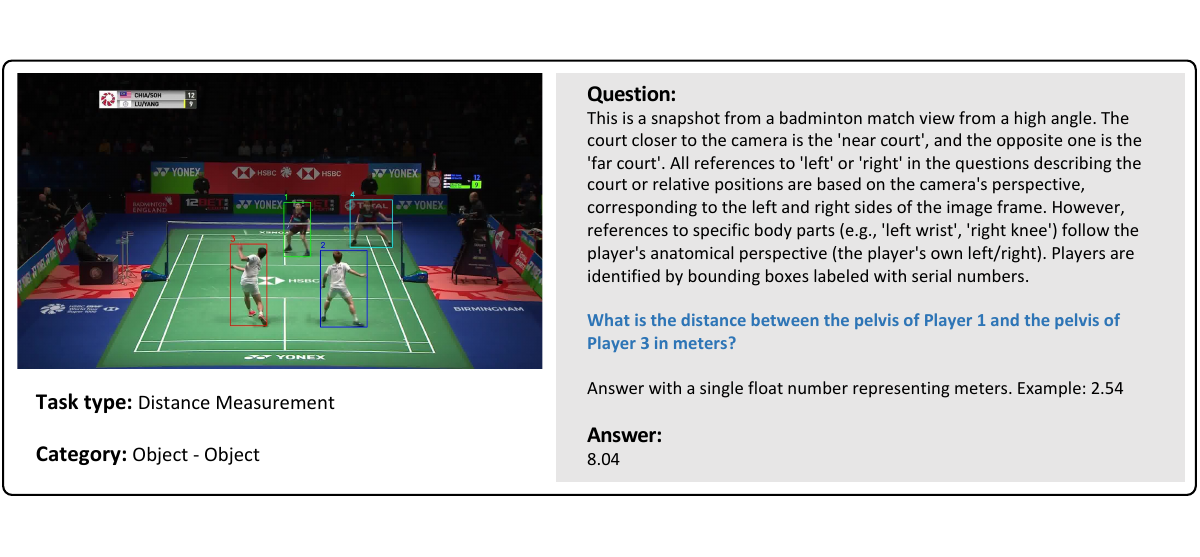}
\end{center}
\vspace{-10mm}

\begin{center}
\includegraphics[width=\linewidth,height=0.25\textheight,keepaspectratio]{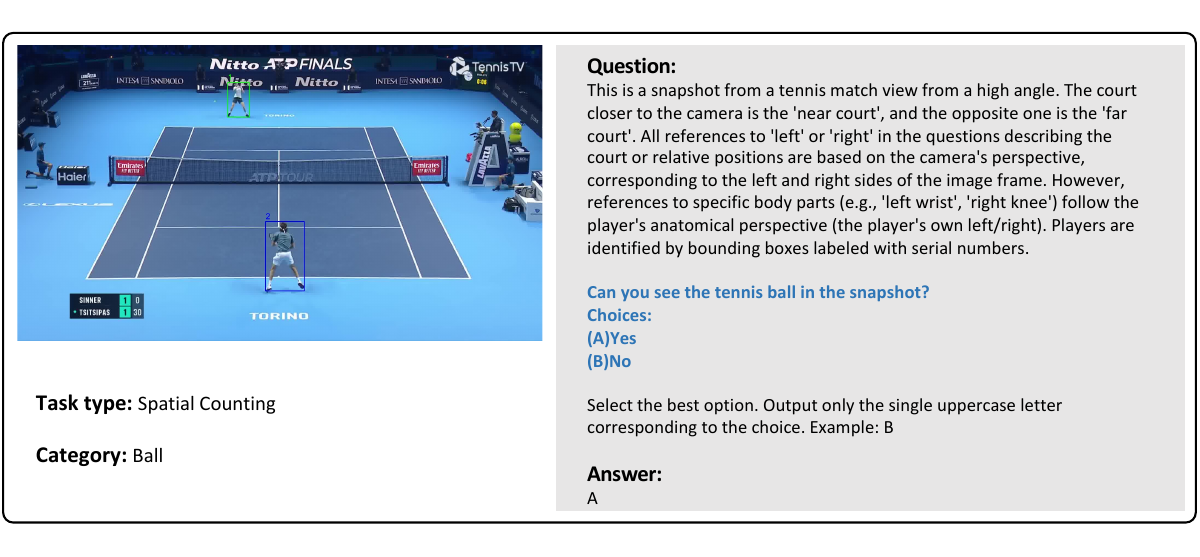}
\end{center}
\vspace{-10mm}

\begin{center}
\includegraphics[width=\linewidth,height=0.25\textheight,keepaspectratio]{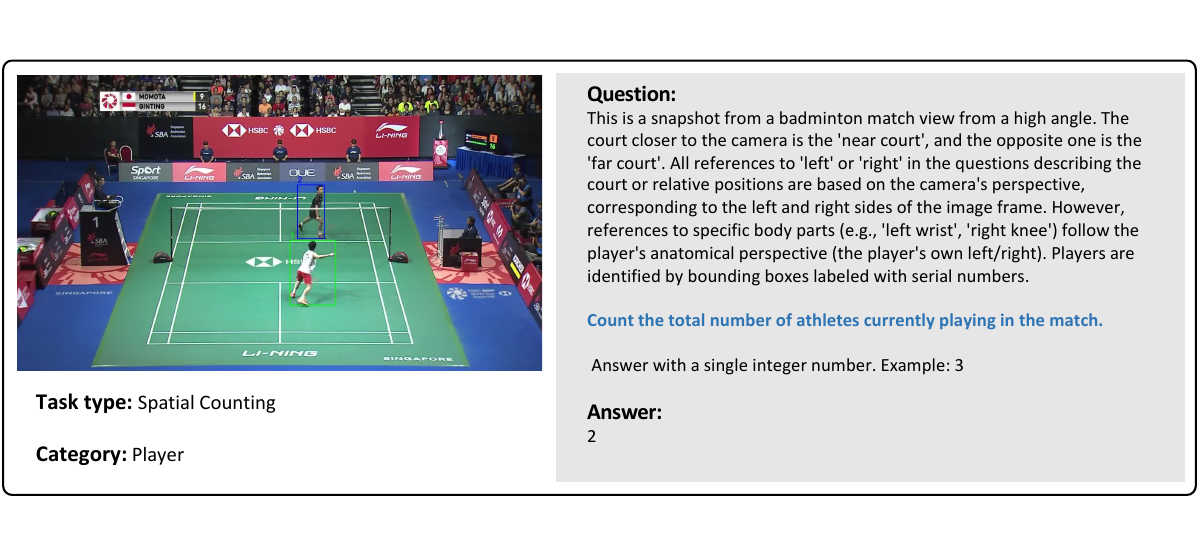}
\end{center}
\vspace{-10mm}

\begin{center}
\includegraphics[width=\linewidth,height=0.25\textheight,keepaspectratio]{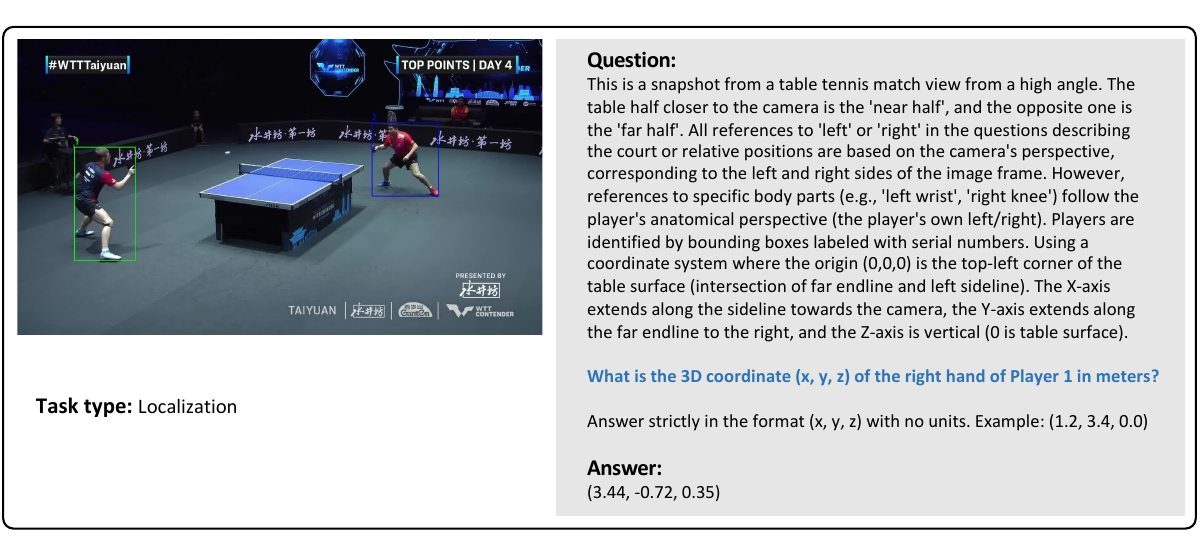}
\end{center}

\begin{center}
\includegraphics[width=\linewidth,height=0.25\textheight,keepaspectratio]{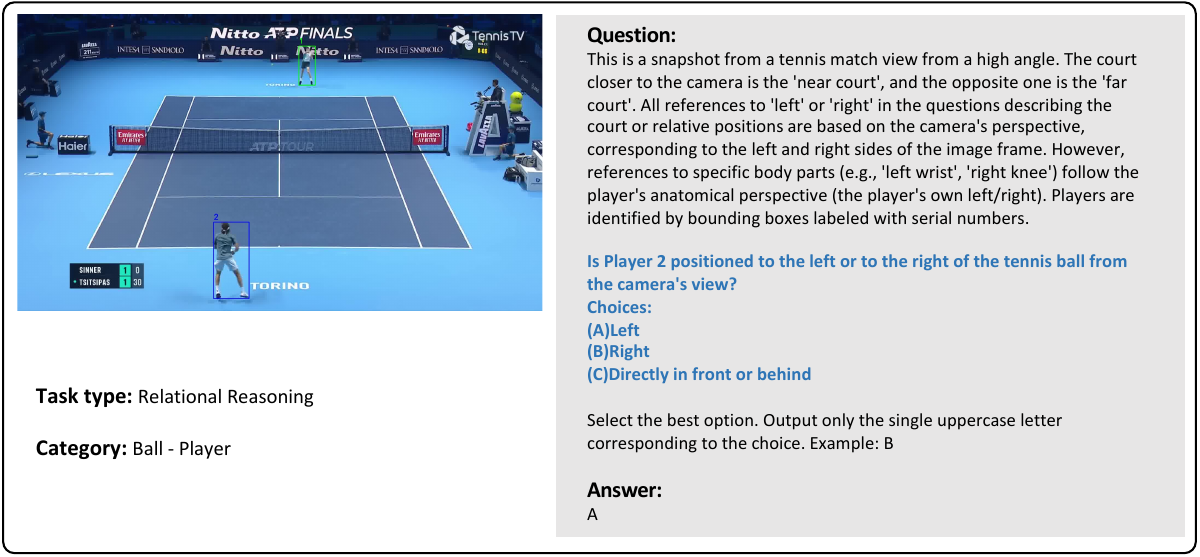}
\end{center}
\vspace{-5mm}

\begin{center}
\includegraphics[width=\linewidth,height=0.25\textheight,keepaspectratio]{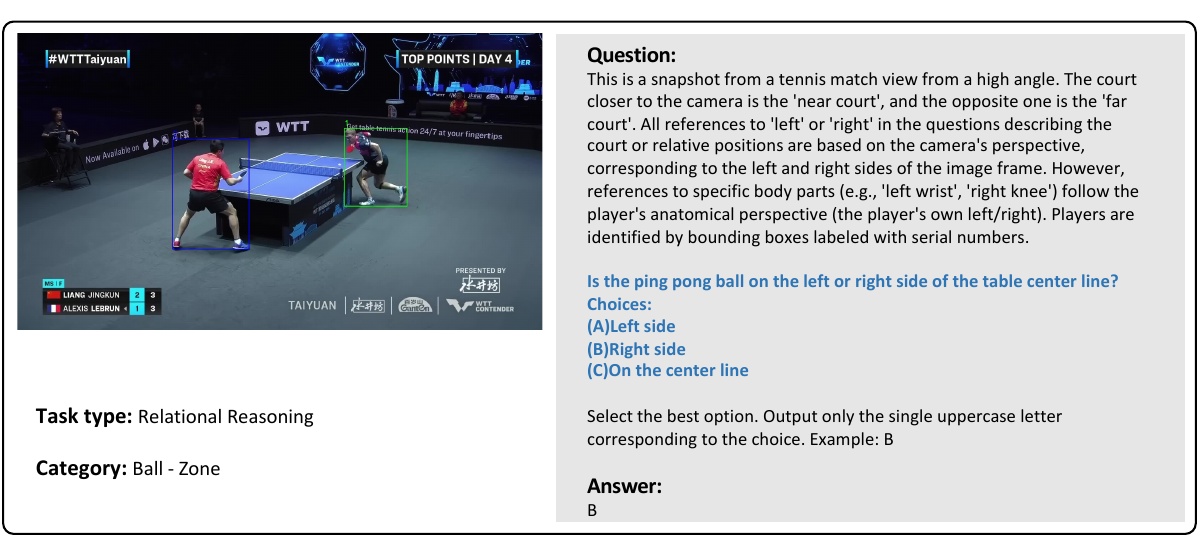}
\end{center}
\vspace{-10mm}

\begin{center}
\includegraphics[width=\linewidth,height=0.25\textheight,keepaspectratio]{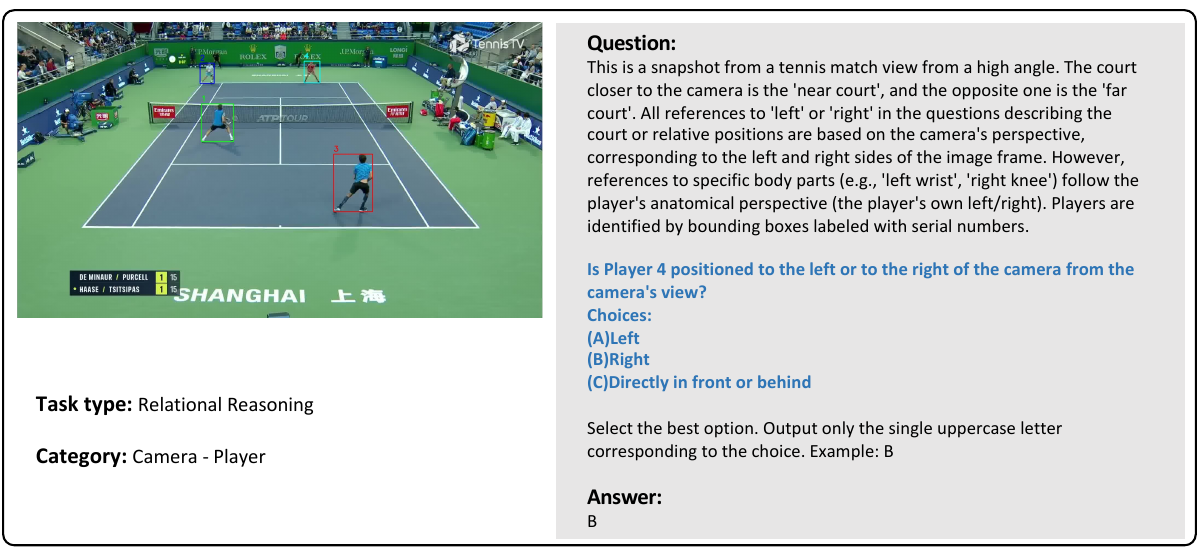}
\end{center}
\vspace{-10mm}

\begin{center}
\includegraphics[width=\linewidth,height=0.25\textheight,keepaspectratio]{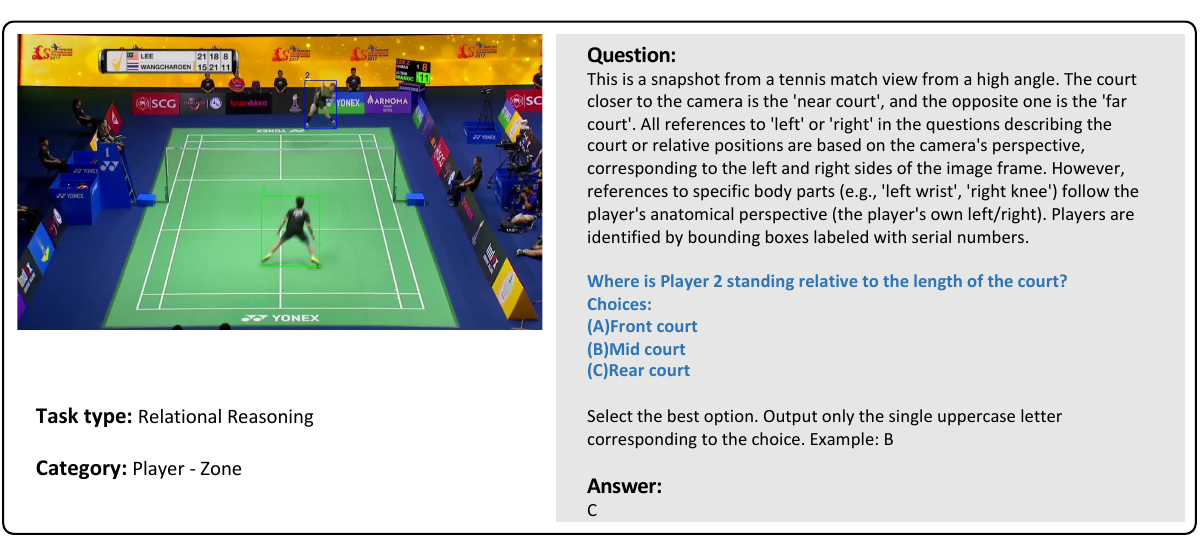}
\end{center}
\vspace{-10mm}

\begin{center}
\includegraphics[width=\linewidth,height=0.25\textheight,keepaspectratio]{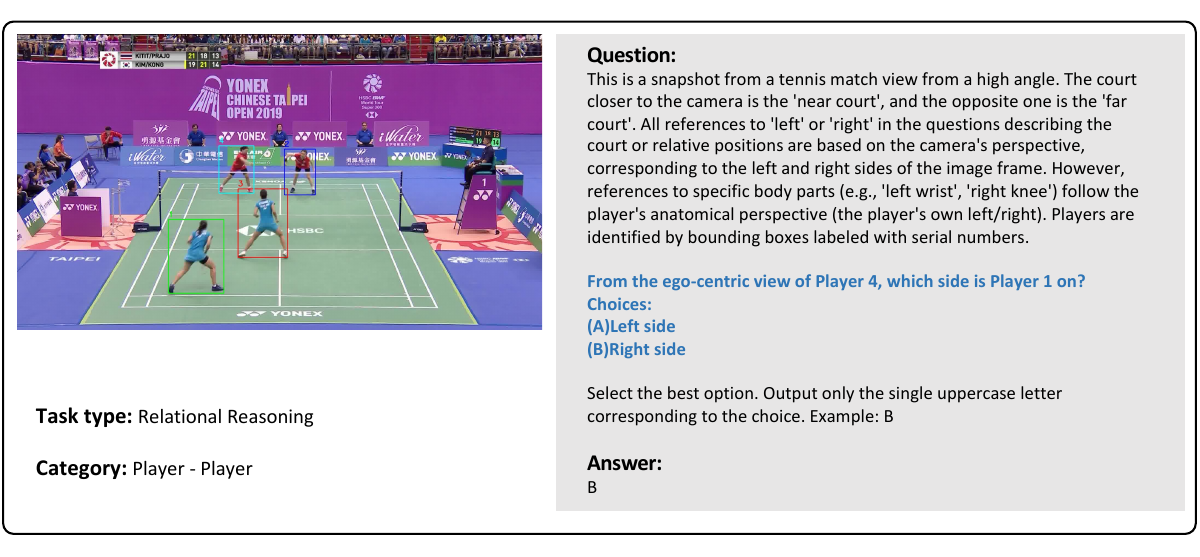}
\end{center}
\vspace{-5mm}

\begin{center}
\includegraphics[width=\linewidth,height=0.25\textheight,keepaspectratio]{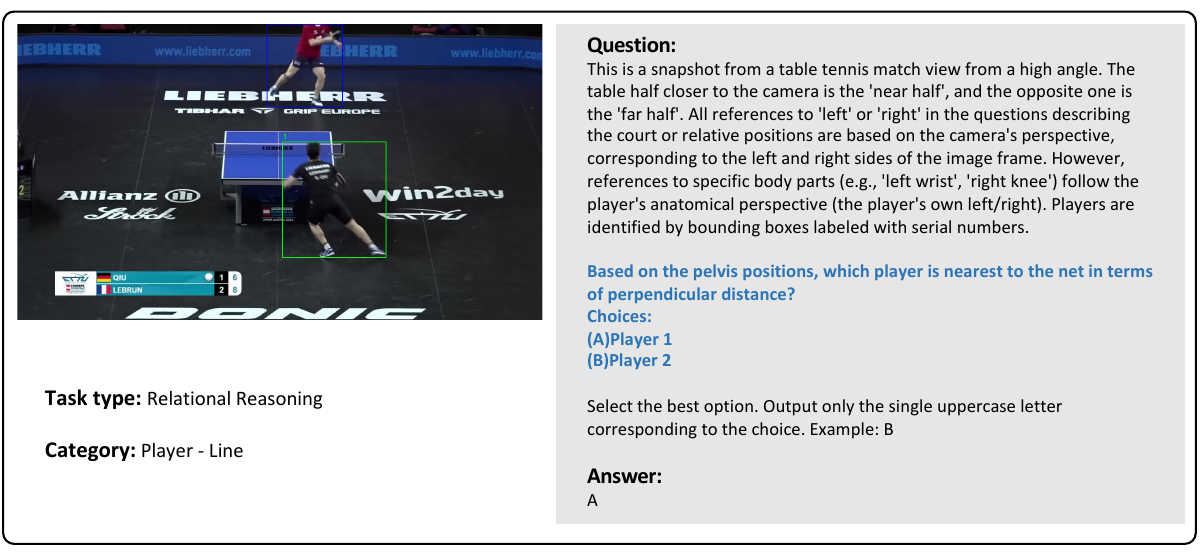}
\end{center}
\vspace{-5mm}

\subsection{Human Review}
As described in the main text, all QA pairs in CourtSI-Bench undergo a final round of manual verification. Any pair flagged by either annotator is removed to ensure annotation quality and consistency.
After this filtering process, we resample the remaining questions according to task categories and per-sport distribution to maintain a balanced benchmark. The final CourtSI-Bench contains 3,686 QA pairs, selected from 4,356 raw samples.
Most discarded instances are due to ambiguous questions or the resampling procedure. For example, because players occupy a non-negligible physical width, certain left/right spatial relationships can be inherently unclear, leading to potential ambiguity for evaluation.
In CourtSI, we introduce task-specific thresholds for each sport to mitigate this issue.

\subsection{Data Distribution}
\label{sec:supp-data-dist}
\begin{table}[htbp]
\centering
\caption{Detailed Data Distribution. B, T, and TT denote badminton, tennis, and table tennis, respectively.}
\label{tab:category_distribution}
\scriptsize
{
\begin{tabular}{cl|@{\hspace{2pt}}c@{\hspace{2pt}}
                @{\hspace{2pt}}c@{\hspace{2pt}}
                @{\hspace{2pt}}c@{\hspace{2pt}}
                @{\hspace{2pt}}c@{\hspace{2pt}}|
                @{\hspace{2pt}}c@{\hspace{2pt}}
                @{\hspace{2pt}}c@{\hspace{2pt}}
                @{\hspace{2pt}}c@{\hspace{2pt}}
                @{\hspace{2pt}}c@{\hspace{2pt}}}
\toprule
\multicolumn{2}{c|}{\multirow{2}{*}{\textbf{Category Name}}}
& \multicolumn{4}{c|}{\textbf{CourtSI-Bench}} 
& \multicolumn{4}{c}{\textbf{CourtSI}} \\
\cmidrule(lr){3-6}\cmidrule(lr){7-10}
& & \textbf{Count} & \textbf{B} & \textbf{T} & \textbf{TT}
  & \textbf{Count} & \textbf{B} & \textbf{T} & \textbf{TT} \\
\midrule

\multirow{4}{*}{\makecell{Distance \\ Measurement}}
  & Camera-Object  & 277 & 27.80\% & 35.74\% & 36.46\% & 75,783  & 33.93\% & 24.41\% & 41.66\% \\
  & Height         & 229 & 23.58\% & 37.12\% & 39.30\% & 51,154  & 31.43\% & 25.00\% & 43.56\% \\
  & Object-Line    & 317 & 24.61\% & 44.16\% & 31.23\% & 102,054 & 31.09\% & 25.20\% & 43.71\% \\
  & Object-Object  & 663 & 25.34\% & 41.18\% & 33.48\% & 178,878 & 29.95\% & 25.55\% & 44.50\% \\
\midrule

\multirow{2}{*}{\makecell{Spatial \\ Counting}}
  & Ball           & 28  & 25.00\% & 42.86\% & 32.14\% & 23,015  & 31.02\% & 25.10\% & 43.88\% \\
  & Player         & 34  & 23.53\% & 32.35\% & 44.12\% & 22,897  & 31.03\% & 25.33\% & 43.63\% \\
\midrule

\multirow{1}{*}{Localization}
  & -              & 368 & 31.25\% & 39.67\% & 29.08\% & 101,698 & 31.02\% & 25.24\% & 43.74\% \\
\midrule

\multirow{6}{*}{\makecell{Relational \\ Reasoning}}
  & Ball-Zone      & 255 & 25.88\% & 32.16\% & 41.96\% & 61,997  & 20.05\% & 22.35\% & 57.60\% \\
  & Ball-Player    & 297 & 24.24\% & 40.40\% & 35.35\% & 72,232  & 24.92\% & 27.26\% & 47.82\% \\
  & Camera-Player  & 248 & 25.40\% & 43.15\% & 31.45\% & 58,280  & 26.54\% & 28.61\% & 44.85\% \\
  & Player-Zone    & 82  & 51.22\% & 48.78\% & -       & 28,769  & 55.31\% & 44.69\% & -       \\
  & Player-Player  & 393 & 44.27\% & 28.24\% & 27.48\% & 104,961 & 45.42\% & 17.83\% & 36.75\% \\
  & Player-Line    & 495 & 32.32\% & 40.00\% & 27.68\% & 127,223 & 31.00\% & 25.37\% & 43.63\% \\
\bottomrule
\end{tabular}
}
\label{tab:supp-detailed-distribution}
\end{table}
In \cref{tab:supp-detailed-distribution}, we present the detailed sample counts and per-sport percentages for both CourtSI-Bench and CourtSI. Overall, the data distribution in CourtSI-Bench across different sports is relatively balanced.

Notably, the Player-Zone subtask under Relational Reasoning primarily describes a player’s relative position within the near or far zones of the court. Since table tennis players do not stand on the table surface itself, these instances are excluded to maintain the validity and consistency of the annotations.

\section{Experiment Details}
\label{sec:supp-exp}

\subsection{Evaluation on CourtSI-Bench Details}
For data parsing, we use the Qwen3-8B model to extract answers from the original model outputs. The detailed prompt is provided below.
\begin{prompt}
Please extract the answer from the following VLM response. Only provide the answer without any explanation. If the answer cannot be found in the VLM response, please output ``None''.

We will give you the original question and the VLM response. Please strictly follow the format to answer.

<Original Question>: \{\textit{Question}\}

<VLM Response>: \{\textit{VLM Answer}\}

<Extracted Answer>:
\end{prompt}

For human evaluation, evaluators are provided with the image and the corresponding question through an interactive panel. The information provided to evaluators is identical to that given to the VLMs. In addition, the court size of each sport is provided as a reference.

In the localization task, the output is represented as 3D coordinates, which prevents the use of T-MRA for computing relative distance error. Therefore, we adopt a binary accuracy metric with a smooth threshold of $30\mathrm{cm}$. 
Notably, this threshold is greater than the combined 3D distance threshold, $15 \times \sqrt{3}$.
If the 3D localization error exceeds this threshold, the prediction is assigned an accuracy of 0; otherwise, it is assigned an accuracy of 1.

\begin{figure}[t]
    \centering
    \includegraphics[width=\linewidth]{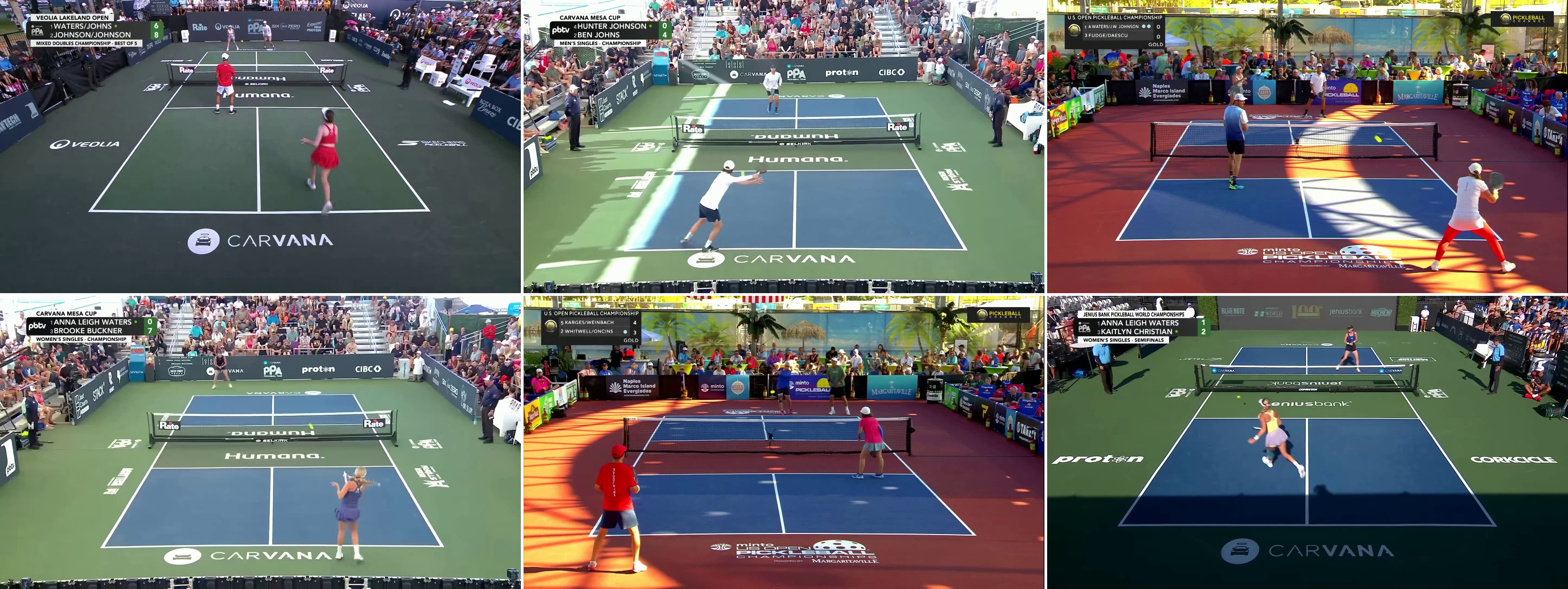}
    \caption{Court-Ext examples.}
    \label{fig:court-ext}
\end{figure}

\subsection{In-depth Error Analysis Details}
We specifically select the object-object and object-line subtasks in CourtSI-Bench, as these tasks are particularly susceptible to perspective ambiguity affecting the target entities.
For the target subjects in each QA pair, we compute the ratio between their 3D distance (in meters) and their 2D projected distance (in pixels) as a quantitative metric. A higher ratio indicates that two subjects are far apart in 3D space but appear close in the 2D image plane, reflecting stronger perspective distortion.
Note that not all reasoning processes are equally affected by perspective. Nevertheless, this ratio serves as a useful proxy for measuring the degree of perspective-induced ambiguity.
Overall, model performance degrades as the perspective effect increases. In certain points of \cref{fig:adv-curve}, performance temporarily improves, likely because some reasoning instances are less dependent on perspective cues, as previously mentioned.

\subsection{CourtSI-Ext Details}
\label{sec:supp-courtsi-ext}
Similar to RacketVision, we collect pickleball videos from YouTube. To ensure a relatively balanced image distribution, the collected videos include both men's and women's matches, as well as singles and doubles matches. All videos are provided at 1080p resolution and 25 fps, consistent with the CourtSI source data.
Examples are illustrated in \cref{fig:court-ext}.

\subsection{Spatial-aware Commentary Generation Details}
We extract spatial relationships from distance measurement cases in CourtSI-Bench, as these cases can be reliably evaluated using ground-truth numerical distance annotations. 
We use the following prompt to instruct baseline models to incorporate spatial relationships into sports commentary generation.
\begin{prompt}
    You are a professional live sports commentator with 3D spatial intelligence.

    Primary Task: Generate a vivid and natural sports commentary describing the scene.

    Spatial Intelligence Task: You need to smoothly incorporate the following spatial relationship into the commentary.

    Here is the relationship: \{Relationship description of objects\}.
    
    Remember to make the commentary engaging and informative, as if you are describing the scene to an audience watching the game live. Avoid directly answering the question or explaining the reasoning steps; instead, focus on creating a rich and immersive commentary that captures the essence of the moment. But the commentary must include the numerical value from the relationship.
\end{prompt}

We sample 100 commentaries from all distance measurement cases for the user study involving three volunteers. For each instance, volunteers compare the outputs from the fine-tuned model and the base model, and evaluate their relative quality in terms of linguistic quality and spatial awareness.
Linguistic quality refers to the overall fluency, expressiveness, and natural integration of spatial information into the commentary. Spatial awareness refers to the correctness and proper use of the numerical distance information described in the spatial relationship.

\section{Ethical Considerations}
\label{sec:supp-ethical-cons}
Our dataset is primarily constructed from images derived from RacketVision, which collects data from publicly available YouTube videos of international net sports games.
We only use image frames for research purposes and do not attempt to identify individuals or infer sensitive personal attributes.
We encourage responsible use of our dataset and recommend that future users comply with ethical research standards in RacketVision.

\end{CJK*}
\end{document}